\documentclass[11pt]{article}

\usepackage[preprint]{acl}

\usepackage{times}
\usepackage{latexsym}

\usepackage{tcolorbox}
\usepackage{multirow}
\usepackage{amsmath}
\usepackage{lineno}
\usepackage{booktabs}
\usepackage{multirow}
\usepackage[table]{xcolor}
\usepackage{colortbl}
\usepackage{tcolorbox}
\tcbuselibrary{skins, breakable}

\definecolor{bestbg}{RGB}{255,235,235}
\definecolor{audiotext}{RGB}{235,245,255}
\definecolor{anytoany}{RGB}{247,245,255}

\usepackage[T1]{fontenc}
\usepackage[utf8]{inputenc}

\usepackage{microtype}

\usepackage{inconsolata}

\usepackage{graphicx}

\title{MM-FinEval: A Multi-Task Multimodal Benchmark\\ for Real-World Financial Forecasting}

\author{
    Dong Shu$^{1}$, Yanguang Liu$^{2}$, Huopu Zhang$^{3}$, Saisai Hu$^{4}$, Haiyan Zhao$^{2}$, Hekun Huang$^{2}$, Mengnan Du$^{5}$ \\
    \\
    $^{1}$Northwestern University \quad
    $^{2}$New Jersey Institute of Technology \quad
    $^{3}$Georgia Institute of Technology \\
    $^{4}$Pace University \quad
    $^{5}$The Chinese University of Hong Kong, Shenzhen
}

\begin{document}
\maketitle
\begin{abstract}
  Financial forecasting from earnings conference calls requires models to reason over complex corporate disclosures, market expectations, and subtle communication signals. However, existing financial benchmarks are often limited to unimodal inputs or single-task settings, making it difficult to evaluate whether multimodal large language models (LLMs) can support real-world financial analysis. In this paper, we introduce MM-FinEval, a novel benchmark designed to evaluate multimodal LLMs across multiple financial tasks. MM-FinEval spans a diverse timeline from 2019 to 2022. The entire proposed dataset contains 2,045 S\&P 500 conference earning calls as inputs and 12 financial task labels as outputs. Each input contains three modalities: a word-to-word text transcript of the earning call, the corresponding presentation slides used during the call, and the entire audio recording. To establish a rigorous evaluation framework, we analyze 19 baseline models across three distinct model categories: Image-Text, Audio-Text, and Any-to-Any configurations. We observe that small-size Any-to-Any models processing all three modalities achieve strong performance, even when compared against larger proprietary models restricted to two-modality inputs. This indicates that our tri-modal dataset design introduces useful, non-redundant information. These results validate that text, audio, and visual data serve as important, complementary signals that mimic the decision-making process of expert human analysts. The MM-FinEval dataset is available at \url{https://github.com/Tizzzzy/MM-FinEval}.
  % \url{https://anonymous.4open.science/r/MM-FinEval-B62B/}.

\end{abstract}

\section{Introduction}

\begin{figure*}
    \centering
    \includegraphics[width=1\linewidth]{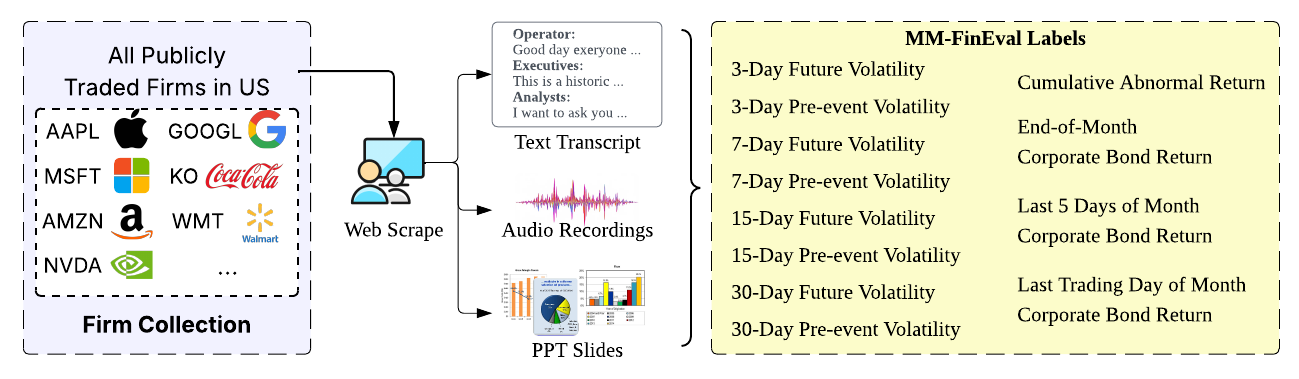}
    \caption{MM-FinEval Construction Pipeline. The dataset construction follows a three-stage process: assembling a foundational collection of publicly traded US firms, systematically web-scraping synchronized tri-modal data (text transcripts, audio recordings, and slides), and annotating the multimodal inputs with 12 financial task labels.}
    \label{fig:pipeline}
    % \vspace{-10pt}
\end{figure*}

Earnings conference calls are important events for public companies, serving as a primary channel for management to communicate financial performance, strategic outlooks, and operational updates to the market. The quality of a company's earnings conference call is directly linked to the next-day stock market performance. In practice, financial analysts rely heavily on these events, extracting  signals from company presentations and Q\&A sessions to analyze corporate health and forecast future performance. Concurrently, there is a rapidly growing trend of leveraging LLMs to analyze financial markets and automate data processing \cite{wu2023bloomberggpt, li2023large, huang2023finbert, yang2023fingpt}. If LLMs can be trained to successfully capture the dense, multi-layered information presented during these calls, financial institutions could benefit from an automated and scalable AI finance analysis pipeline that has historically required substantial human analysis.

Despite the rapid advancement of LLMs, existing financial models remain constrained by their reliance on uni-modal architectures and their tendency to focus on narrow, specific financial tasks \cite{he2025cross, zhu2025post, liang2025does, konstantinidis2024finllama, liu2025fin}. A major reason for this limitation is that the majority of current financial datasets are heavily skewed toward a single modality, such as pure text or numerical time-series data \cite{chen2021finqa, lai2025sec, shah2023trillion, dong2024fnspid}. Furthermore, even when recent datasets incorporate multimodal elements beyond text, they are typically restricted to a specific task \cite{shu2025fincall, li2020maec, luo2025finmme, shu2025finchart}. This combination of unimodal and task-specific training data severely restricts financial LLMs to learn from additional contextual information, ultimately limiting their broader applicability in complex, real-world financial environments.

To bridge this gap, we propose MM-FinEval, a large-scale, multi-modal, and multi-label financial benchmark dataset designed specifically for earnings call analysis. Spanning a period from 2019 to 2022, the dataset encompasses a total of 2045 data points from S\&P 500 companies, with 450 from 2019, 273 from 2020, 499 from 2021, and 823 from 2022. Each data point integrates three modalities: word-to-word text transcripts of the earnings call, the corresponding raw earnings call audio, and the accompanying presentation slide decks. Furthermore, MM-FinEval is annotated with 12 labels, providing a benchmark to evaluate the performance of multimodal financial models. Through an extensive experiment with 19 state-of-the-art models, we found that:

\begin{itemize}
    \item Current LLMs exhibit a severe performance gap between market return and volatility forecasting.
    \item Integrating all three modalities provides performance advantage, proving that non-redundant visual and audio information are essential for replicating expert financial reasoning.
    \item Model performance on volatility tasks aligns with the financial theory of information half-life. As the forecasting horizon expands from 3 to 30 days, the earnings signal is rapidly diluted by unpredictable market factors.  
\end{itemize}

\section{Related Work}

\subsection{Earnings Conference Call Datasets}

Earnings conference calls are important events for public companies because the quality of the information disclosed during these presentations and Q\&A sessions is directly linked to next-day stock market performance. Recently, researchers have increasingly deployed machine learning models to systematically interpret the information in these calls \cite{sawhney2020voltage, sawhney2020risk, mathur2022docfin, sawhney2020multimodal}, aiming to uncover quantitative investment signals and making money. This effort to uncover quantitative investment signals has motivated the community to publish specialized datasets focusing on earnings conference calls. For example, ECTSum \cite{mukherjee-etal-2022-ectsum} provides benchmarks for bullet-point summarization of highly verbose transcripts. FinNLI \cite{magomere-etal-2025-finnli} is developed to assess logical deduction. SubjECTive-QA \cite{pardawala2024subjective} measures the subjectivity of Q\&A responses across conference earning calls. Recent large text datasets like STRUX \cite{lu2024strux} and DEC \cite{yu2025same} created for LLM-driven investment decision-making.

However, these existing datasets remain constrained by their reliance on unimodal architectures, do not include the acoustic and visual signals inherent in the actual conference call. EC \cite{qin2019you} which was the first dataset to combine textual transcripts with raw audio recordings to predict financial risk. Building upon this, MAEC \cite{li2020maec} scaled the available conference call data and optimized to predict stock volatility. Datsets such as Earnings-21 \cite{del2021earnings} and ConEC \cite{huang-etal-2024-conec} focus heavily on the acoustic challenges of the domain, serving primarily as benchmarks for automatic speech recognition rather than broader financial reasoning. FinCall-Surprise \cite{shu2025fincall}, which provides conference call text transcripts, audio, and slides, is optimized exclusively for predicting earnings surprises. Despite the success of these existing multi-modal conference call datasets, they are limited to a specific financial task. This severely restricts the ability of financial LLMs to learn from additional contextual information, directly motivating the need for multi-label multi-modal benchmarks.

\begin{table*}[ht]
\centering
\scalebox{1}{
\begin{tabular}{l|ccc|ccc|ccc|c}
\toprule
& \multicolumn{3}{c|}{Transcript (words)} & \multicolumn{3}{c|}{Slide (pages)} & \multicolumn{3}{c|}{Audio (sec)} &  \\ 
\cmidrule(lr){2-4} \cmidrule(lr){5-7} \cmidrule(lr){8-10}
Year & Mini & Max & Avg. & Mini & Max & Avg. & Mini & Max & Avg. & Total \\ 
\midrule
2019 &  2477 & 15421 & 8837.90 & 1 & 156 & 25.97 & 1096.07 & 8870.97 & 3642.66 & 450 \\
2020 & 2533 & 21894 & 9389.90 & 1 & 170 & 27.89 & 1404.06 & 7585.31 & 3859.25 & 273 \\
2021 & 2784 & 15343 & 8998.85 & 1 & 113 & 27.47 & 1197.95 & 11938.89 & 3720.29 & 499 \\
2022 & 2986 & 17717 & 8842.89 & 2 & 252 & 30.73 & 234.72 & 31393.57 & 3770.45 & 823 \\
\bottomrule
\end{tabular}}
\caption{Data statistics of the MM-FinEval. The dataset spans four years (2019–2022), with each conference call containing three synchronized modalities: text transcripts, presentation slides, and audio recordings. We report the minimum, maximum, and average values for each modality. Transcript length is measured in words, slides in pages, and audio in seconds. The last column shows how much data we have for each year.}
\label{tab:data_analysis}
% \vspace{-10pt}
\end{table*}

\section{MM-FinEval Construction}

The construction of MM-FinEval dataset follows a three-stage pipeline designed to collect, synchronize, and annotate data from multiple sources. 
% Our primary objective is to construct a robust benchmark that integrates the textual, audio, and visual information of conference earnings calls. 
As illustrated in Figure \ref{fig:pipeline}, the first stage involves gathering a set of liquid public firms in the United States. To ensure data quality, we restrict the sample to companies with a market capitalization exceeding \$1 billion and an average daily trading volume above \$50 million. This initial screening process yielded a pool of over 4,000 companies.

In the second stage, we acquire data across the three modalities for each firm’s earnings conference calls. We systematically web-scrape these contents from official company websites. The collected text transcripts are highly structured, featuring clear speaker identifiers (operators, executives, and analysts) alongside their corresponding statements. Because most publicly traded firms host one earnings call per fiscal quarter, we ensure temporal consistency by manually aligning all data sources according to their specific reporting dates. Furthermore, we cross-reference the conference call titles to verify that the transcript, audio recording, and presentation slides all originate from the exact same event, ensuring synchronization across our multimodal dataset.

\subsection{Multi-Label Preliminaries}

MM-FinEval benchmark includes 12 labels designed to predict a company's cumulative abnormal return, stock volatility, and corporate bond returns based on its quarterly earnings call. The complete labels are shown on the right side of Figure \ref{fig:pipeline}.

\paragraph{\textbf{Cumulative Abnormal Return.}}
Cumulative abnormal return (CAR) measures the market-adjusted price reaction around an earnings call event \cite{hirshleifer2009driven, duan2018learning}. For each firm $i$ and event date $t$, we first compute the abnormal return by subtracting a benchmark market return from the firm’s realized return:
\begin{equation}
AR_{i,t} = R_{i,t} - R_{m,t},
\end{equation}
where $R_{i,t}$ denotes the stock return of firm $i$ on day $t$, and $R_{m,t}$ denotes the corresponding market return. We then aggregate abnormal returns over an event window $[t_1, t_2]$:
\begin{equation}
CAR_{i,[t_1,t_2]} = \sum_{t=t_1}^{t_2} AR_{i,t}.
\end{equation}
CAR reflects whether the information revealed in the earnings call leads to a positive or negative market reassessment of the firm. We include CAR-based labels over multiple post-call windows to evaluate whether models can infer short-term and medium-term directional market reactions from multimodal earnings-call information.

\paragraph{\textbf{Volatility.}}
Volatility measures stock-price fluctuations over a finite period and is commonly used as a measure of risk \cite{kogan2009predicting}. In our study, the period is defined relative to the earnings conference call using pre-event and post-event windows of different lengths. Following Kogan et al. \cite{kogan2009predicting}, volatility from day $t-\tau$ to day $t$ is defined as
% \begin{equation}
% v_{[t-\tau,t]}
% =
% \ln\left(
% \sqrt{
% \frac{\sum_{i=0}^{\tau}(r_{t-i}-\bar{r})^2}{\tau}
% }
% \right),
% \end{equation}

\begin{equation}
v_{[t-\tau,t]}
=
\sqrt{
\frac{\sum_{i=0}^{\tau}(r_{t-i}-\bar{r})^2}{\tau}
},
\end{equation}

where $r_t$ is the return on day $t$, $\bar{r}$ is the mean return over the period of day $t - \tau$ to day $t$. $r_t=P_t/P_{t-1}-1$ is the return price, with $P_t$ denoting the closing price on day $t$. While CAR captures the direction of the market response, volatility captures stock-price instability regardless of whether returns are positive or negative. The labels \texttt{fut\_3d}--\texttt{fut\_30d} and \texttt{past\_3d}--\texttt{past\_30d} measure post-event and pre-event volatility, respectively, over windows of 3, 7, 15, and 30 calendar days.

\paragraph{\textbf{Corporate Bond Return.}}
Corporate bond returns capture the monthly performance of a firm's outstanding bonds and provide a debt-market measure of investors' responses to information disclosed during earnings calls. The bond return data are observed at the monthly level and are matched to conference-call observations using the firm's stock ticker, recorded as \texttt{COMPANY\_SYMBOL}, and the bond trading month, recorded in \texttt{DATE}. Specifically, monthly returns are computed as the percentage change in the bond's dirty price (clean price plus accrued interest), adjusted for coupon payments received during the holding period, following the standard methodology in the corporate bond literature \cite{bessembinder2008measuring, jostova2013momentum}. To account for differences in bond trading frequency and liquidity, we consider three complementary return measures that differ only in their definition of the month-end price. \textit{RET\_EOM} is calculated using the last observed transaction price during the calendar month, regardless of the trading date. \textit{RET\_LDM} uses the transaction price on the last trading day of the month and is missing if the bond does not trade on that day. \textit{RET\_L5M} uses the last transaction price observed within the final five trading days of the month, providing a balance between month-end pricing and sample coverage.

\section{Data Analysis}

\begin{figure*}
    \centering
    \includegraphics[width=0.8\linewidth]{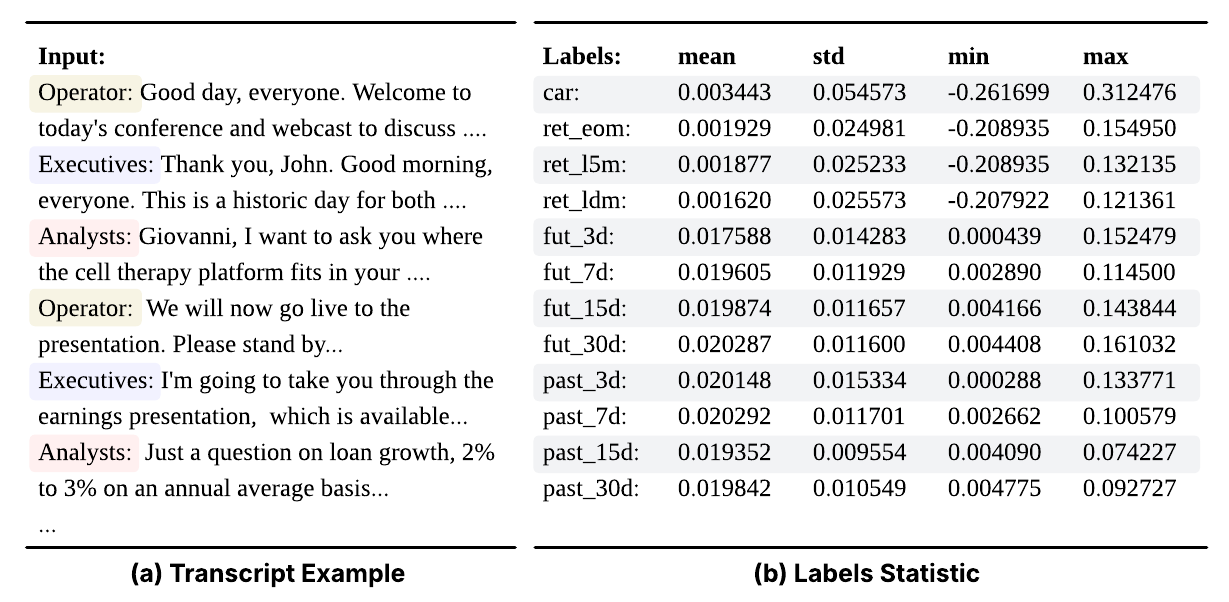}
    \caption{Transcript Structure and Label Statistics.}
    \label{fig:transcript_example}
\end{figure*}

Table \ref{tab:data_analysis} presents a detailed statistical analysis of our dataset, summarizing its scale and multimodal structure. MM-FinEval has a total of 2045 data spanning from 2019 to 2022. On average, an individual data features a text transcript of approximately 9,000 words, accompanied by a 27-page presentation slide deck and an audio recording spanning roughly 3,700 seconds. As illustrated in Figure \ref{fig:transcript_example}a, the content of these calls is consistently structured around three primary speaker roles: Operators, Executives, and Analysts. Operators typically manage the flow of the event by opening and closing the call. Executives, such as the CEO and CFO, deliver the company’s financial results and strategic outlook. Finally, Analysts representing investment firms and financial institutions ask questions to obtain more information into the company’s operational and financial performance. More data analysis including our label statistic is in Appendix \ref{app:data_analysis}.

\section{Benchmark Settings}

\subsection{Baseline LLMs}

To establish a comprehensive performance benchmark for MM-FinEval, we selected 19 state-of-the-art open- and closed-source models. These models were categorized into three architectural groups based on their input capabilities.

The first group comprises Large Vision-Language Models (LVLMs), which process both image and text inputs. This group includes the Qwen 3.5 series (spanning 2B to 35B parameters) \cite{qwen3.5}, the Qwen 3.6 series (27B and 35B) \cite{qwen36_35b_a3b}, Gemma 3n \cite{team2024gemma}, and the Mistral 3 series (3B, 8B, and 14B) \cite{jiang2023mistral7b}. It also includes proprietary models: GPT-5.4, GPT-5.5, Claude 4.6 Sonnet, and Claude 4.8 Opus. The second group consists of Text-Audio Language Models, which process both audio and text inputs, represented in our study by Voxtral 24B \cite{liu2025voxtral} and Moss 8B \cite{mossaudio2026}. Finally, the third group features Any-to-Any Models, which are natively equipped to process all modalities, text, audio, and images, simultaneously. This includes Gemma 4 E2B and Gemma 4 E4B \cite{gemmateam2026gemma4}.

\subsection{Benchmark Input Design}

To systematically evaluate the performance of the baseline models, we designed a customized input and prompting strategy tailored to each modality. The full set of prompts used in the experiments is provided in Appendix \ref{app:prompts}.

\paragraph{\textbf{Image-Text Modality: }}
Because the selected baseline models represent the current state-of-the-art, they all support extended context windows exceeding 128K tokens. Given that the longest text transcript in our dataset is 21,894 words, we input the raw transcripts directly without applying any preprocessing steps, such as textual summarization.
However, the visual data required structural adaptation. The original presentation slides are collected in PDF format. Since most models do not natively support PDF input, we converted each page into a separate image. To accommodate the image-count limits inherent to many of these models, we implemented a random sampling strategy. For each presentation, we randomly selected five images from the slide deck, explicitly excluding the first and last pages, which typically contain non-substantive content such as title cards or concluding remarks. In cases where a presentation contained five or fewer substantive slides, all available slides were used. The final input sequence was structured as: ``\{Images\} \{Instruction\} \{Text Transcript\} \{Answer Format\}''. Although this specific benchmark experiment restricts the visual input to five sampled images, the complete PDF files are included in the public dataset release.

\paragraph{\textbf{Audio-Text Modality: }}
We structured the input sequence as: ``\{Audio\} \{Instruction\} \{Text Transcript\} \{Answer Format\}''.

\paragraph{\textbf{Image-Audio-Text Modality: }}

To ensure a fair comparison across all baseline models, we maintained the exact input constraints established in the previous setups. We used the same randomly selected slides as the Image-Text setup. The final input sequence for this comprehensive modality was structured as: ``\{Images\} \{Audio\} \{Instruction\} \{Text Transcript\} \{Answer Format\}''.

\begin{table*}[ht]
\centering
\renewcommand{\arraystretch}{1.4}
\resizebox{\textwidth}{!}{
\begin{tabular}{ll ccccccccccccccc cc cc}
\toprule
\toprule
\multirow{2}{*}{Label} & \multirow{2}{*}{Metric} & \multicolumn{15}{c}{\textit{Image-Text}} & \multicolumn{2}{c}{\textit{Audio-Text}} & \multicolumn{2}{c}{\textit{Any-to-Any}} \\ \cmidrule(lr){3-17} \cmidrule(lr){18-19} \cmidrule(lr){20-21}
 & & Qw3.5-2B & Qw3.5-4B & Qw3.5-9B & Qw3.5-27B & Qw3.5-35B & Qw3.6-27B & Qw3.6-35B & Gm3n & Mis3-3B & Mis3-8B & Mis3-14B & GPT-5.4 & GPT-5.5 & Cld-4.6 & Cld-4.8 & Vox-24B & Moss & Gm4-E2B & Gm4-E4B \\ \hline
\multirow{2}{*}{car} & MAE $\downarrow$ & 0.0511 & 0.0498 & 0.0499 & 0.0534 & 0.0470 & 0.0505 & 0.0478 & 0.0740 & 0.0599 & 0.0540 & 0.0546 & 0.0493 & 0.0526 & 0.0493 & \cellcolor{bestbg}\textbf{0.0458} & \cellcolor{audiotext}\textbf{0.0429} & 0.1079 & \cellcolor{anytoany}\textbf{0.0418} & 0.0437 \\
 & DA $\uparrow$ & 21.76\% & 41.61\% & 48.07\% & 50.56\% & 53.11\% & 51.64\% & 51.10\% & \cellcolor{bestbg}\textbf{55.88\%} & 46.11\% & 50.66\% & 51.98\% & 51.83\% & 53.45\% & 53.69\% & 41.09\% & 20.79\% & \cellcolor{audiotext}\textbf{37.60\%} & \cellcolor{anytoany}\textbf{55.17\%} & 53.20\% \\ \hline
\multirow{2}{*}{ret\_eom} & MAE $\downarrow$ & 0.0341 & 0.0198 & 0.0210 & 0.0302 & 0.0199 & 0.0228 & 0.0196 & 0.0346 & 0.0184 & 0.0184 & 0.0184 & 0.0182 & \cellcolor{bestbg}\textbf{0.0144} & 0.0179 & 0.0177 & \cellcolor{audiotext}\textbf{0.0224} & 3.4277 & 0.0297 & \cellcolor{anytoany}\textbf{0.0248} \\
 & DA $\uparrow$ & 37.85\% & 25.28\% & 51.00\% & 56.23\% & 55.55\% & 56.77\% & 57.11\% & 56.65\% & 53.55\% & 56.82\% & 56.33\% & 59.49\% & 71.46\% & 71.92\% & \cellcolor{bestbg}\textbf{72.41\%} & 21.71\% & \cellcolor{audiotext}\textbf{37.16\%} & \cellcolor{anytoany}\textbf{60.59\%} & 60.10\% \\ \hline
\multirow{2}{*}{ret\_l5m} & MAE $\downarrow$ & 0.0316 & 0.0196 & 0.0207 & 0.0221 & 0.0192 & 0.0201 & 0.0192 & 0.0335 & 0.0193 & 0.0246 & 0.0199 & 0.0182 & \cellcolor{bestbg}\textbf{0.0171} & 0.0191 & 0.0192 & \cellcolor{audiotext}\textbf{0.0210} & 0.0748 & 0.0229 & \cellcolor{anytoany}\textbf{0.0216} \\
 & DA $\uparrow$ & 26.50\% & 26.45\% & 52.96\% & 56.67\% & 55.35\% & 56.09\% & 54.03\% & 56.16\% & 5.77\% & 51.25\% & 54.72\% & 59.22\% & 66.01\% & 67.00\% & \cellcolor{bestbg}\textbf{67.49\%} & 23.47\% & \cellcolor{audiotext}\textbf{40.78\%} & 59.11\% & \cellcolor{anytoany}\textbf{61.08\%} \\ \hline
\multirow{2}{*}{ret\_ldm} & MAE $\downarrow$ & 0.0318 & 0.0206 & 0.0206 & 0.0215 & 0.0188 & 0.0195 & 0.0187 & 0.0353 & 0.0198 & 0.0225 & 0.0197 & 0.0183 & \cellcolor{bestbg}\textbf{0.0180} & 0.0188 & 0.0194 & \cellcolor{audiotext}\textbf{0.0231} & 0.1369 & 0.0295 & \cellcolor{anytoany}\textbf{0.0234} \\
 & DA $\uparrow$ & 25.23\% & 34.23\% & 51.10\% & 55.21\% & 54.38\% & 53.74\% & 56.63\% & 55.17\% & 6.45\% & 53.69\% & 54.87\% & 59.22\% & 64.46\% & \cellcolor{bestbg}\textbf{67.98\%} & 66.50\% & 27.09\% & \cellcolor{audiotext}\textbf{39.76\%} & \cellcolor{anytoany}\textbf{62.56\%} & 62.07\% \\ \hline
\multirow{2}{*}{fut\_3d} & MAE $\downarrow$ & 0.0230 & 0.0215 & 0.0262 & 0.0253 & 0.0169 & 0.0214 & 0.0154 & 0.0305 & 0.0261 & 0.0319 & 0.0308 & 0.0252 & 0.0221 & 0.0142 & \cellcolor{bestbg}\textbf{0.0139} & \cellcolor{audiotext}\textbf{0.0192} & 21.7155 & 0.0115 & \cellcolor{anytoany}\textbf{0.0089} \\
~ & DA $\uparrow$ & 47.04\% & 95.11\% & 96.09\% & 96.58\% & 99.41\% & 96.33\% & 99.32\% & 97.54\% & 71.44\% & 97.75\% & 97.26\% & 99.76\% & \cellcolor{bestbg}\textbf{100.00\%} & \cellcolor{bestbg}\textbf{100.00\%} & \cellcolor{bestbg}\textbf{100.00\%} & 35.60\% & \cellcolor{audiotext}\textbf{53.64\%} & 98.03\% & \cellcolor{anytoany}\textbf{99.51\%} \\ \hline
\multirow{2}{*}{fut\_7d} & MAE $\downarrow$ & 0.0235 & 0.0190 & 0.0236 & 0.0227 & 0.0157 & 0.0191 & 0.0156 & 0.0298 & 0.0267 & 0.0749 & 0.0318 & 0.0254 & 0.0242 & 0.0150 & \cellcolor{bestbg}\textbf{0.0106} & \cellcolor{audiotext}\textbf{0.0196} & 0.6668 & 0.0102 & \cellcolor{anytoany}\textbf{0.0087} \\
~ & DA $\uparrow$ & 46.11\% & 95.89\% & 95.84\% & 95.84\% & 99.22\% & 97.11\% & 98.68\% & \cellcolor{bestbg}\textbf{100.00\%} & 66.21\% & 96.43\% & 97.36\% & \cellcolor{bestbg}\textbf{100.00\%} & \cellcolor{bestbg}\textbf{100.00\%} & 99.01\% & \cellcolor{bestbg}\textbf{100.00\%} & 32.08\% & \cellcolor{audiotext}\textbf{53.50\%} & 98.52\% & \cellcolor{anytoany}\textbf{99.51\%} \\ \hline
\multirow{2}{*}{fut\_15d} & MAE $\downarrow$ & 0.0238 & 0.0198 & 0.0241 & 0.0218 & 0.0163 & 0.0205 & 0.0254 & 0.0283 & 0.0242 & 0.1292 & 0.0344 & 0.0265 & 0.0267 & 0.0157 & \cellcolor{bestbg}\textbf{0.0097} & \cellcolor{audiotext}\textbf{0.0194} & 0.6633 & 0.0097 & \cellcolor{anytoany}\textbf{0.0085} \\
~ & DA $\uparrow$ & 46.50\% & 95.89\% & 95.94\% & 98.48\% & 99.27\% & 98.34\% & 99.22\% & 98.03\% & 51.64\% & 94.67\% & 97.26\% & \cellcolor{bestbg}\textbf{100.00\%} & \cellcolor{bestbg}\textbf{100.00\%} & 97.54\% & \cellcolor{bestbg}\textbf{100.00\%} & 24.01\% & \cellcolor{audiotext}\textbf{53.30\%} & 98.52\% & \cellcolor{anytoany}\textbf{99.34\%} \\ \hline
\multirow{2}{*}{fut\_30d} & MAE $\downarrow$ & 0.0246 & 0.0215 & 0.0255 & 0.0227 & \cellcolor{bestbg}\textbf{0.0166} & 0.0574 & 0.1495 & 0.0331 & 0.0257 & 0.2064 & 0.0573 & 0.0287 & 0.0250 & 0.0586 & 0.0191 & \cellcolor{audiotext}\textbf{0.0198} & 0.1338 & 0.0106 & \cellcolor{anytoany}\textbf{0.0086} \\
~ & DA $\uparrow$ & 54.67\% & 95.50\% & 96.72\% & 97.90\% & 98.78\% & 98.09\% & 98.58\% & 98.52\% & 58.68\% & 94.96\% & 97.21\% & \cellcolor{bestbg}\textbf{100.00\%} & \cellcolor{bestbg}\textbf{100.00\%} & 98.03\% & \cellcolor{bestbg}\textbf{100.00\%} & 31.39\% & \cellcolor{audiotext}\textbf{52.81\%} & 98.03\% & \cellcolor{anytoany}\textbf{99.19\%} \\ \hline
\multirow{2}{*}{past\_3d} & MAE $\downarrow$ & 0.0206 & 0.0213 & 0.0331 & 0.0229 & 0.0177 & 0.0180 & 0.0219 & 0.0397 & 0.0393 & 0.1095 & 0.0337 & 0.0150 & 0.0131 & \cellcolor{bestbg}\textbf{0.0118} & 0.0123 & \cellcolor{audiotext}\textbf{0.0229} & 0.1823 & 0.0172 & \cellcolor{anytoany}\textbf{0.0152} \\
~ & DA $\uparrow$ & 4.01\% & 81.08\% & 95.26\% & 99.17\% & 95.45\% & 98.88\% & 84.16\% & 93.60\% & 53.64\% & 95.31\% & 99.32\% & \cellcolor{bestbg}\textbf{100.00\%} & \cellcolor{bestbg}\textbf{100.00\%} & \cellcolor{bestbg}\textbf{100.00\%} & \cellcolor{bestbg}\textbf{100.00\%} & 47.33\% & \cellcolor{audiotext}\textbf{56.87\%} & \cellcolor{anytoany}\textbf{94.58\%} & 55.17\% \\ \hline
\multirow{2}{*}{past\_7d} & MAE $\downarrow$ & 0.0205 & 0.0206 & 0.0347 & 0.0214 & 0.0153 & 0.0138 & 0.0288 & 0.0427 & 0.0509 & 0.2498 & 0.0414 & 0.0142 & 0.0166 & 0.0143 & \cellcolor{bestbg}\textbf{0.0114} & \cellcolor{audiotext}\textbf{0.0226} & 0.7153 & 0.0207 & \cellcolor{anytoany}\textbf{0.0140} \\
~ & DA $\uparrow$ & 3.42\% & 81.12\% & 95.06\% & 99.66\% & 98.04\% & 99.32\% & 89.19\% & 96.06\% & 48.95\% & 95.35\% & 99.27\% & \cellcolor{bestbg}\textbf{100.00\%} & \cellcolor{bestbg}\textbf{100.00\%} & \cellcolor{bestbg}\textbf{100.00\%} & \cellcolor{bestbg}\textbf{100.00\%} & 46.70\% & \cellcolor{audiotext}\textbf{56.48\%} & \cellcolor{anytoany}\textbf{91.63\%} & 57.14\% \\ \hline
\multirow{2}{*}{past\_15d} & MAE $\downarrow$ & 0.0209 & 0.0206 & 0.0344 & 0.0214 & 0.0150 & 0.0157 & 0.0668 & 0.0482 & 0.0484 & 0.2786 & 0.0546 & 0.0138 & 0.0177 & 0.0340 & \cellcolor{bestbg}\textbf{0.0118} & \cellcolor{audiotext}\textbf{0.0216} & 0.7578 & 0.0403 & \cellcolor{anytoany}\textbf{0.0125} \\
~ & DA $\uparrow$ & 8.85\% & 81.17\% & 95.06\% & 99.66\% & 99.12\% & 99.46\% & 89.63\% & 93.43\% & 32.57\% & 95.84\% & 99.36\% & \cellcolor{bestbg}\textbf{100.00\%} & \cellcolor{bestbg}\textbf{100.00\%} & \cellcolor{bestbg}\textbf{100.00\%} & \cellcolor{bestbg}\textbf{100.00\%} & 47.48\% & \cellcolor{audiotext}\textbf{57.21\%} & \cellcolor{anytoany}\textbf{92.61\%} & 63.05\% \\ \hline
\multirow{2}{*}{past\_30d} & MAE $\downarrow$ & 0.0228 & 0.0212 & 0.0374 & 0.0356 & \cellcolor{bestbg}\textbf{0.0186} & 0.1199 & 0.2120 & 0.0605 & 0.0564 & 0.2872 & 0.1753 & 0.0245 & 0.0243 & 0.1068 & 0.0602 & \cellcolor{audiotext}\textbf{0.0241} & 1.1258 & 0.0742 & \cellcolor{anytoany}\textbf{0.0120} \\
~ & DA $\uparrow$ & 15.99\% & 78.78\% & 94.72\% & 99.61\% & 99.46\% & 99.32\% & 94.72\% & 95.07\% & 35.94\% & 95.75\% & 98.24\% & \cellcolor{bestbg}\textbf{100.00\%} & \cellcolor{bestbg}\textbf{100.00\%} & \cellcolor{bestbg}\textbf{100.00\%} & \cellcolor{bestbg}\textbf{100.00\%} & 50.71\% & \cellcolor{audiotext}\textbf{56.48\%} & \cellcolor{anytoany}\textbf{91.63\%} & 69.46\% \\
\bottomrule
\bottomrule
\end{tabular}}
\caption{Overall Performance of 19 Baseline Models on MM-FinEval. The evaluation reports averaged Mean Absolute Error (MAE) and Directional Accuracy (DA) across 12 financial tasks spanning the 2019–2022 period. For each row, the highest score for each metric is highlighted in a distinct color corresponding to the model type.}
% \vspace{-10pt}
\label{tab:benchmark_result}
\end{table*}

\subsection{Benchmark Metrics}
% \textcolor{red}{TODO}

Because all of our labels are related to stock returns or volatility, the ground truth data consists of continuous decimal values. Therefore, we frame the prediction as a regression task. In this context, Mean Absolute Error (MAE) serves as the default metric for evaluating absolute model performance on our benchmark. For a dataset of $n$ samples, MAE is defined as:

\begin{equation}
    \text{MAE} = \frac{1}{n} \sum_{i=1}^{n} \vert{}y_i - \hat{y}_i\vert{},
\end{equation}

where $y_i$ represents the ground-truth numerical label and $\hat{y}_i$ denotes the model's prediction. However, given the inherent complexities of financial forecasting, relying on a single evaluation metric is insufficient to capture a model's true reasoning capabilities. Therefore, we also employ the Pearson Correlation Coefficient ($r$), formulated as:

\begin{equation}
    r = \frac{\sum_{i=1}^{n} (y_i - \bar{y})(\hat{y}_i - \bar{\hat{y}})}{\sqrt{\sum_{i=1}^{n} (y_i - \bar{y})^2 \sum_{i=1}^{n} (\hat{y}_i - \bar{\hat{y}})^2}}.
\end{equation}

This metric evaluates the model's capacity to correctly identify the relative magnitude of market reactions across a broad cross-section of companies. Finally, we report Directional Accuracy (DA). In financial markets, accurately forecasting the direction of a movement is a important signal for investment strategies. DA measures the proportion of predictions that correctly anticipate the mathematical sign of the actual outcome, expressed as:

\begin{equation}
    \text{DA} = \frac{1}{n} \sum_{i=1}^{n} \mathbf{1}(\text{sgn}(y_i) = \text{sgn}(\hat{y}_i)),
\end{equation}

where $\mathbf{1}$ is the indicator function and $\text{sgn}$ extracts the positive or negative sign of the value.

\subsection{Implementation Details}

All experiments were conducted on two NVIDIA A100 GPUs, each with 80GB of memory. For all baseline models, we maintained their official repository configurations and loaded them with bfloat16 precision to optimize computational efficiency. Beyond this precision setting, no other modifications were made to the models’ default parameters, ensuring a fair and reproducible comparison across all baseline evaluations.

\section{Benchmark Results}

\subsection{The Gap Between Returns and Volatility}
\label{sec:returns_and_volatility}

As shown in Table \ref{tab:benchmark_result}, an initial analysis of the benchmark results reveals a large gap in the ability of LLMs to predict market returns versus market volatility. Across the majority of baseline models, Directional Accuracy (DA) for return-based tasks (car, ret\_eom, ret\_l5m, ret\_ldm) is relatively poor, frequently ranging between 40\% and 60\%. Only few models achieve above 70\% (e.g., Claude-4.8 on ret\_eom). Because DA evaluates only the binary sign of the prediction, a score hovering near 50\% indicates that these models only marginally outperform random guessing. In contrast, the majority of models achieve a DA over 90\%, and occasionally reach 100\%, when predicting volatility. Exceptions are rare, with only a few small-size models, such as Qwen3.5-2B, falling below 50\%. However, as illustrated by the label statistics in Figure \ref{fig:transcript_example}b, it is common knowledge in the financial domain that volatility cannot be negative. Therefore, despite achieving high DA scores on volatility-based tasks, small-size LLMs still fail to preserve a basic understanding of fundamental constraints in financial domains.

This performance gap is further illuminated by evaluating the Mean Absolute Error (MAE). For instance, high-performing models (e.g., Claude, Gemma-4) achieve MAEs between 0.041 and 0.045 on the Cumulative Abnormal Return (car) task. As shown in Figure \ref{fig:transcript_example}b, return-based tasks are inherently noisy: their ground truth means hover near zero while exhibiting high variance. Consequently, this MAE is large compared to car ground truth mean of 0.003443 and representing approximately 77\% of the dataset's standard deviation. In contrast, volatility tasks demonstrate a tighter error distribution. On tasks like fut\_15d, top models achieve MAEs as low as 0.0085 to 0.0097, which represent a much smaller fraction of the ground truth mean (0.019874) and standard deviation (0.011657). 
% Yet, while MAEs on volatility are smaller, the models' failure to respect basic non-negativity bounds reveals it lacks structural financial knowledge. This underscores the importance of utilizing diverse evaluation metrics to accurately assess LLM capabilities in financial tasks. Ultimately, current LLMs fail to demonstrate reliable quantitative reasoning for either continuous return or volatility forecasting. 
We present a detailed breakdown of each model's annual MAE performance in Table~\ref{tab:MAE_trend}, capturing the granular trends over each consecutive year from 2019-2022.

\begin{figure*}[ht]
    \centering
    \includegraphics[width=0.8\linewidth]{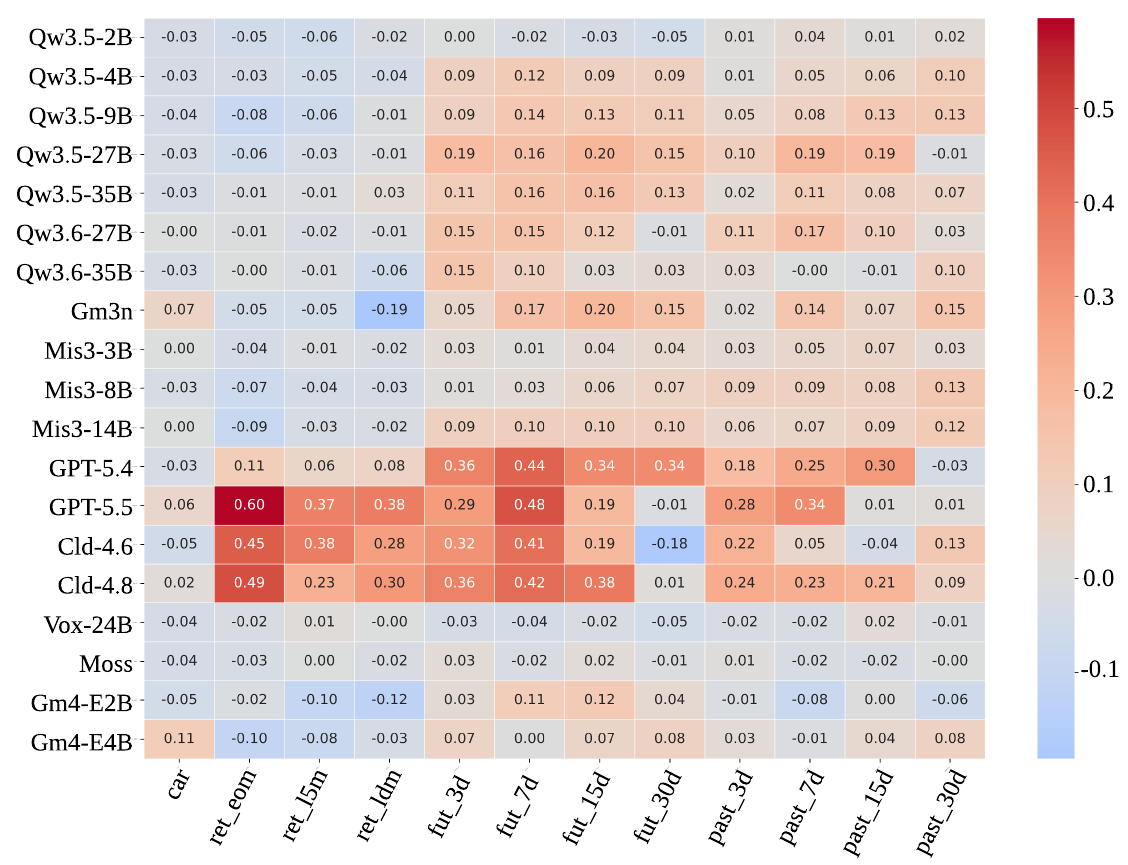}
    \caption{Overall Pearson Correlation Coefficients of Baseline Models. The heatmap illustrates the models' capacity to recognize linear trends in financial data.}
    \label{fig:pearson}
    % \vspace{-8pt}
\end{figure*}

\subsection{Inconsistent Scaling Laws and the Proprietary Advantage}

Within specific model architectures and tasks, our benchmark in Table \ref{tab:benchmark_result} reveals a clear scaling law where increased parameter count directly correlates with enhanced financial reasoning capabilities. This trend is particularly pronounced within the Qwen 3.5 and 3.6 model family. As the parameter size scales from 2 billion to 35 billion, we observe a consistent, upward trajectory in overall Directional Accuracy across both return-based and volatility tasks. 

However, this scaling law breaks down on the MAE metric. Rather than exhibiting a consistent decrease as parameter counts grow, the models' MAE changes unpredictably across tasks. For example, within the Qwen 3.5 family on the Cumulative Abnormal Return task, the 27B model yields an MAE of 0.0534, performing worse than its much smaller 4B (0.0498) and 9B (0.0499) counterparts. Similar inconsistencies appear across other model families. For instance, the Mistral-3B model achieves a lower MAE (0.0261) on the fut\_3d task than both the Mistral-8B (0.0319) and Mistral-14B (0.0308). These findings suggest that while larger models have a better ability to predict the market movement direction, they do not have the knowledge to accurately estimate the magnitude of those movements.

Beyond model family scaling, the benchmark results highlight a performance gap between open-source and proprietary models. Within the Image-Text modality, the GPT and Claude model families consistently outperform their open-source counterparts. Leading proprietary models, such as GPT-5.5 and Claude-4.8, achieve the highest tier of Directional Accuracy across the majority of tasks. Specifically, they reach DA scores of approximately 72\% on ret\_eom, and 67\% on both ret\_l5m and ret\_ldm. These results outperform the near-random guessing baseline (approximately 50\%) in return based tasks. Furthermore, their 100\% DA on volatility tasks suggests that these advanced proprietary models have a basic understanding of the non-negativity constraints inherent to financial volatility.

\begin{figure*}[ht]
    \centering
    \includegraphics[width=1\linewidth]{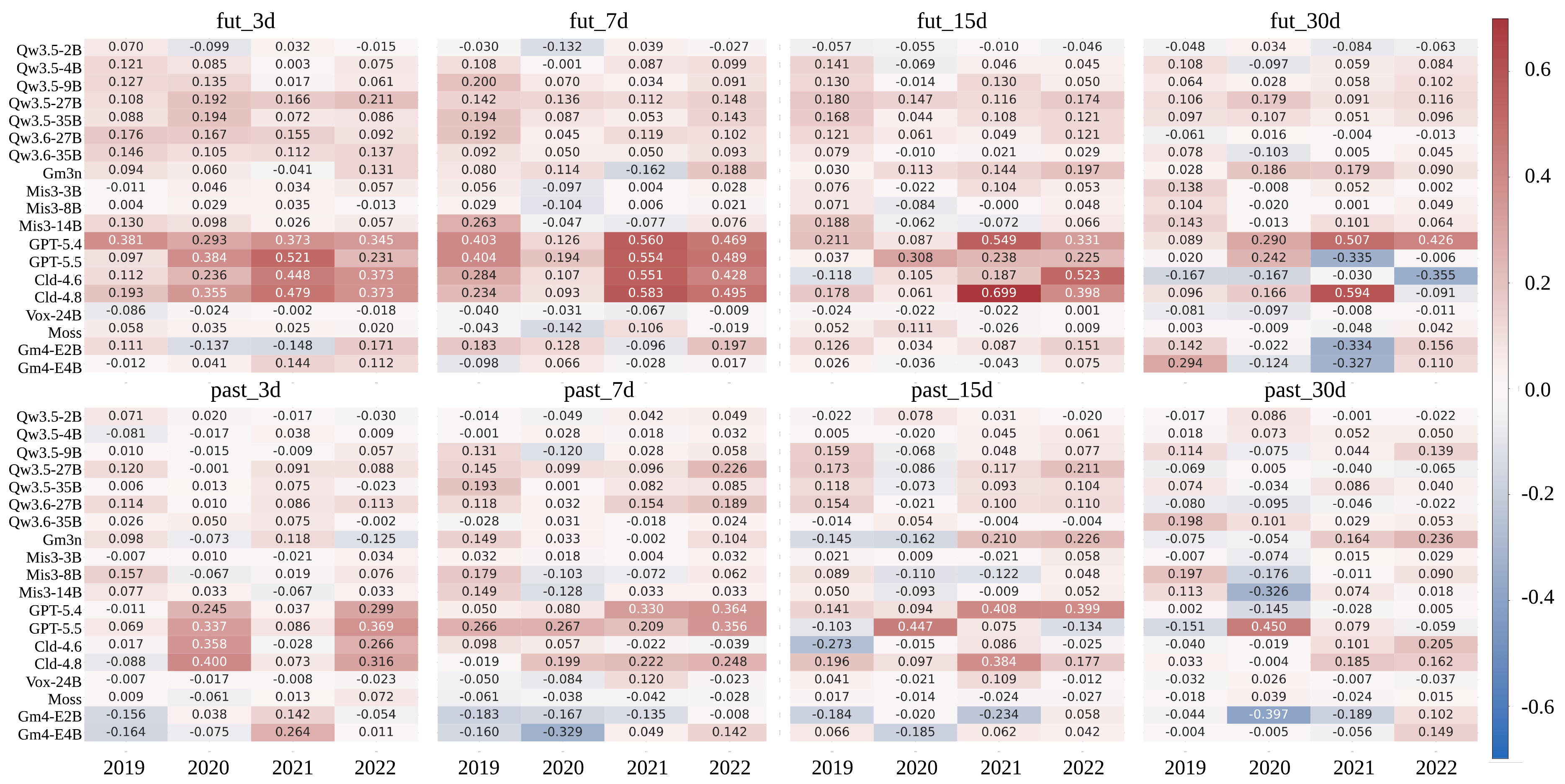}
    \caption{Annual Pearson Correlation Trends (2019–2022) on Volatility Tasks. The heatmap illustrates how models' ability to capture linear relationships shifts over time.}
    \label{fig:pearson_trend_vol}
    % \vspace{-8pt}
\end{figure*}

\subsection{The Advantage of Any-to-Any Multimodal Integration}

As shown in Table \ref{tab:benchmark_result}, our benchmark results show the advantage of integrating all three data modalities: text, audio, and visual. The Any-to-Any models evaluated in this study, specifically Gemma 4 E2B and Gemma 4 E4B, perform well on both return-based and volatility tasks. Despite their small size architectures (Gemma 4 E2B is a 5B parameter model utilizing only 2B active parameters, while Gemma 4 E4B is an 8B model with 4B active parameters) both achieve a Directional Accuracy exceeding 60\% across ret\_eom, ret\_l5m, and ret\_ldm. This performance is better than much larger Image-Text modality baselines, such as the 35B parameter Qwen 3.6 model, and even proprietary GPT 5.4 model. Furthermore, on the Cumulative Abnormal Return task, both Gemma 4 E2B (0.0418) and Gemma 4 E4B (0.0437) achieve lower MAEs than the highly capable Claude-4.8 model (0.0458).

This unexpected performance strongly validates that textual, visual, and audio data streams provide distinct, complementary financial signals rather than redundant information. In real-world financial markets, institutional analysts do not evaluate corporate disclosures solely based on the conference call transcript. They synthesize what management is presenting via visual slides, the literal lexical content of the transcript, and, critically, how that content is delivered, extracting subtle acoustic cues from vocal tone, pacing, and inflection. To accurately predict market reactions, LLMs must develop a similarly holistic comprehension. The superior performance of these Any-to-Any models shows that equipping LLMs with this complete spectrum of human communication is essential for replicating expert-level financial reasoning.

\subsection{Conservative Versus Aggressive Prediction in Audio-Text Models}

Another interesting observation in Table \ref{tab:benchmark_result} is that, in Audio-Text columns, Voxtral consistently achieves a superior MAE across the entire benchmark, whereas Moss performs better on DA. A closer examination reveals that Voxtral exhibits highly conservative prediction behaviors, frequently generating its output close to zero. By minimizing the absolute magnitude of its predictions, Voxtral successfully avoids absolute error penalties, resulting in a relatively lower MAE. However, this zero-anchored conservatism prevents the model from capturing meaningful market momentum, leading to exceptionally poor DA scores. Conversely, Moss adopts a highly aggressive predictive strategy. It frequently generates big numerical prediction, and is more likely to capture the correct directional sign.

\begin{figure*}[ht]
    \centering
    \includegraphics[width=0.95\linewidth]{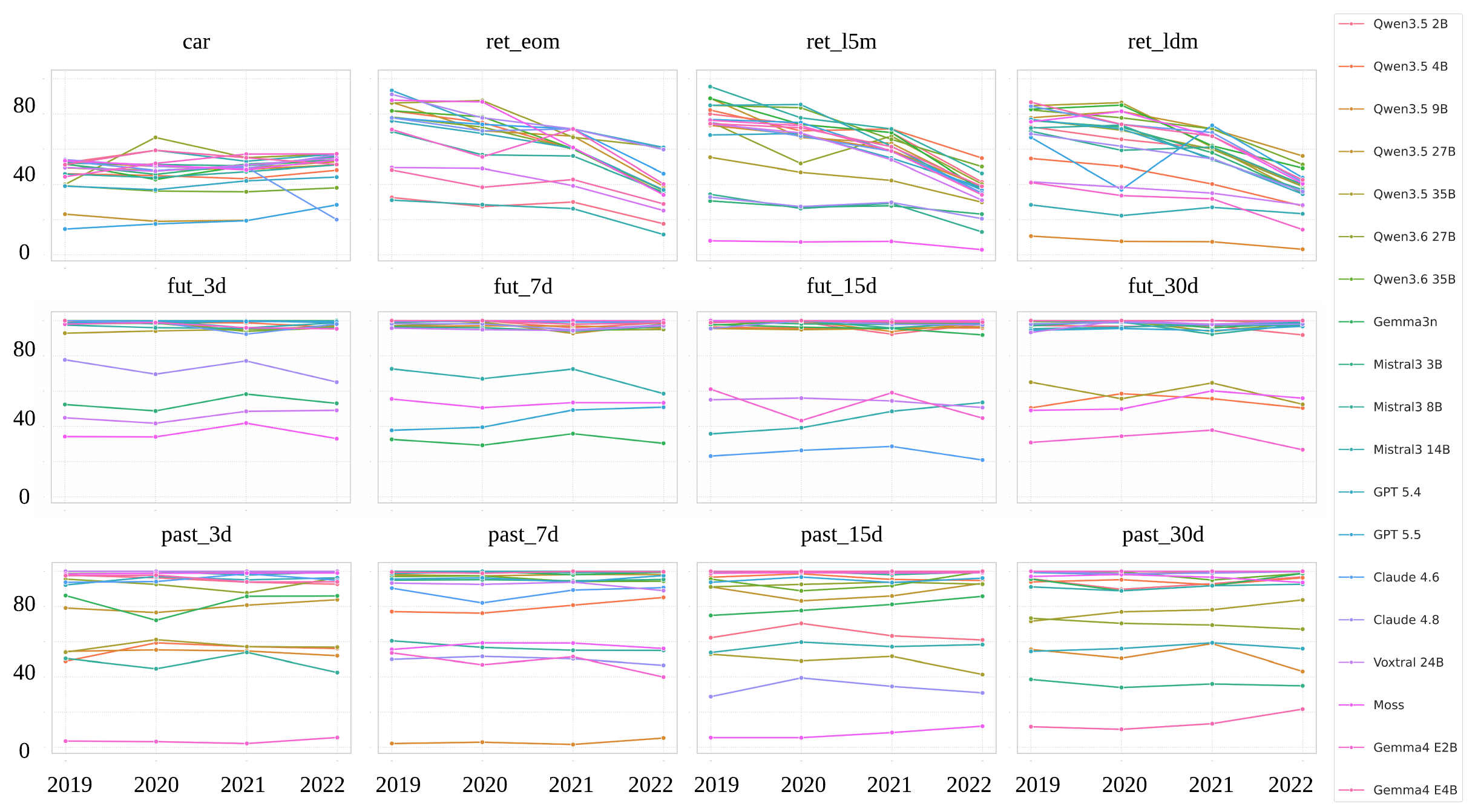}
    \caption{Annual Directional Accuracy (DA) Trends Across MM-FinEval (2019–2022). Each line represents an individual model's performance over the four-year period, illustrating how predictive accuracy shifts across different financial tasks.}
    % \vspace{-10pt}
    \label{fig:DA_trend}
\end{figure*}

\subsection{Pearson Correlation}
\label{sec:pearson_correlation}

In addition to MAE and DA, we evaluated the Pearson correlation coefficient to assess the models' capacity to recognize linear trends and relative shifts in the financial data. As illustrated in Figure \ref{fig:pearson}, open-source models generally exhibit low Pearson correlations across all tasks, with the highest scores peaking near 0.20 on the vol\_fut\_15d task (achieved by Qwen3.5 27B and Gemma 3n). Interestingly, proprietary models such as GPT-5.4, GPT-5.5, and the Claude 4 series achieve surprisingly moderate-to-strong positive Pearson correlations on volatility tasks. For instance, GPT-5.4 reaches a correlation of 0.44 on vol\_fut\_7d, and Claude 4.8 reaches 0.42. This results reveals that proprietary architectures can successfully detect relative shifts in market risk.

We also observe a universal failure across all architectures to predict Cumulative Abnormal Return (car). While proprietary models demonstrate strong positive correlations on discrete return metrics, with GPT-5.5 achieving an impressive 0.60 on ret\_eom (End of Month Return). Both open-source and closed-source models exhibit near-zero or even negative correlations on the CAR task. This discrepancy likely stems from the structural complexity of Cumulative Abnormal Return, which isolates the firm-specific stock reaction by mathematically regressing against broader market benchmarks over a specific event window.

% While LLMs can seemingly deduce raw price momentum and general sentiment (as reflected in their success on ret\_eom), they lack the intrinsic quantitative mechanisms required to decouple a company's idiosyncratic performance from broader macroeconomic market noise.

Finally, the correlation data exposes a clear temporal degradation in model performance across volatility prediction horizons. For the proprietary models, predictive capability is highest at the short-term 3-day and 7-day intervals, but gradually decays through the 15-day mark, ultimately collapsing at the 30-day horizon for both post-event and pre-event volatility. This progressive decay strongly aligns with the financial theory of information half-life: the qualitative and quantitative signals extracted directly from an earnings call are most potent in the immediate short term \cite{hafez2021three, molinaro2026earnings}. As the prediction window expands to a full month, the initial catalyst of the earnings signal is rapidly diluted and overwhelmed by unpredictable market factors. This confirms that forecasting long-term variance is inherently unfeasible for current LLM architectures. 

To further validate the information half-life theory, we map the Pearson correlation coefficients for the volatility tasks across the 2019–2022 timeline in Figure \ref{fig:pearson_trend_vol}. This visualization reinforces the theory by revealing a clear temporal degradation as the prediction horizon expands. For instance, when observing the proprietary models across post-event volatility windows, GPT and Claude achieve strong correlations for the 3-day and 7-day intervals (fut\_3d and fut\_7d); however, this predictive capability steadily deteriorates at the 15-day and 30-day marks (fut\_15d and fut\_30d). An isolated exception occurs with Claude 4.8, which exhibits unusually high performance across all post-event volatility tasks specifically in the year 2021. Despite this outlier, the overarching trend of progressive decay strongly aligns with our findings in Figure \ref{fig:pearson}, reaffirming that the predictive signal contained within a corporate disclosure is rapidly diluted by unpredictable market factors over extended temporal windows. More results analysis is in Appendix \ref{app:pearson_temporal}.

\subsection{Temporal Degradation on Direction Accuracy}

To evaluate how model performance changes over time, we analyze the annual Directional Accuracy trends from 2019 to 2022. As illustrated in Figure \ref{fig:DA_trend}, there is a consistent decline in Directional Accuracy from 2019 to 2022 across the return-based tasks (ret\_eom, ret\_l5m, and ret\_ldm). In 2019, several top-performing models tightly cluster near peak accuracies between 70\% and 90\%. However, this performance decreases and drops to a DA range of 30\% to 40\% in 2022. This highlights a lack of temporal generalization in current LLMs. It suggests that post-pandemic market dynamics introduced increased market complexity that reduced the models' predictive capabilities.

% Even more pronounced are the volatility subplots, where almost all models are remain near 0\% DA from 2019 to 2022. The plot visually demonstrates that current models struggle to map corporate call data to risk estimation, resulting in a persistent failure regardless of the annual market environment.

\section{Conclusion}

In this work, we introduced MM-FinEval, a multimodal benchmark for evaluating LLMs across 12 financial tasks using 2,045 S\&P 500 earnings conference calls spanning a 2019–2022 period. The three modalities inputs in MM-FinEval reflect the analysis process of expert human financial analysts, who must synthesize not only what executives disclose but how they present it to execute complex financial reasoning. Our extensive baseline evaluation across Image-Text, Audio-Text, and Any-to-Any configurations exposes the boundary where pure language modeling must transition into multi-sensory, calibrated quantitative reasoning. We hope this dataset serves as a benchmark for the AI and finance communities, supporting future work toward multimodal financial analysis for economic forecasting.

\clearpage


\bibliography{custom}

\clearpage

\appendix

\section{Data Analysis}
\label{app:data_analysis}

In Figure \ref{fig:transcript_example}, we detail the label statistics, including the mean, standard deviation, minimum, and maximum for all 12 financial task labels. An analysis of these distributions explicitly confirms the structural differences between discrete return tasks and continuous volatility tasks.

The return-based tasks (car, ret\_eom, ret\_l5m, and ret\_ldm) exhibit a classic zero-centered financial return distribution. The means for these labels hover barely above zero, ranging from 0.001620 to 0.003443. In contrast, their standard deviations are substantially larger, ranging from 0.024981 to 0.054573. This high variance relative to the mean confirms a notoriously low signal-to-noise ratio inherent to predicting market direction. Furthermore, the minimum and maximum values demonstrate that returns fluctuate wildly in both directions. For instance, the Cumulative Abnormal Return (car) spans from a minimum of -0.261699 to a maximum of 0.312476, illustrating the extreme, bidirectional stochastic shocks that models must account for.

Conversely, the volatility-based tasks display a fundamentally different statistical profile. Most notably, the minimum values across all eight volatility labels are strictly positive (e.g., the lowest absolute minimum is 0.000288 for past\_3d). This directly reflects the hard non-negativity constraint of financial volatility. The means are significantly higher and tightly clustered between 0.017588 and 0.020292. Additionally, the standard deviations (0.009554 to 0.015334) are much smaller relative to their respective means when compared to the return labels. This points to a tighter, more structured distribution characterized by volatility clustering, making it a more autoregressive and stable phenomenon.

These distinct distributional characteristics mathematically explain the performance divergence observed in our benchmark results. Return tasks are heavily obscured by symmetric, zero-centered noise, making them highly resistant to accurate quantitative regression. Volatility tasks, however, present a structured, bounded, and clustered target that is quantitatively easier for models to map, perfectly mirroring established financial econometrics.

\section{Prompt Used in Our Experiments}
\label{app:prompts}

\subsection{CAR Task}

\begin{tcolorbox}[
    enhanced,
    breakable,
    colback=orange!10!white, 
    colframe=blue!5!black,
    arc=2mm,
    boxrule=1pt,
    title={\bfseries Image-Text Models},
    coltitle=white,
    attach boxed title to top left={yshift=-2mm, xshift=3mm},
    boxed title style={enhanced, colback=blue!5!black, colframe=blue!5!black, arc=2mm, boxrule=0pt},
    top=0.5mm, 
    left=1mm, 
    right=1mm, 
    bottom=0.5mm,
]
\setlength{\parskip}{2pt}
\small\vskip8pt

You are an expert quantitative equity analyst. Your task is to predict the stock's Cumulative Abnormal Return (CAR) for a company based on its earnings call transcript and the accompanying presentation slides.

Transcript:

\{text\_transcript\}

Instructions:

1. Predict the Cumulative Abnormal Return (CAR) associated with this earnings event. CAR measures the sum of the differences between the stock's actual return and its expected return over the event window.

2. Provide 1-2 sentences explaining the key driver.

3. Your CAR prediction must be represented as a decimal.

**Answer format**

Rationale = [[ 1-2 sentences rationale ]]

Prediction = [[ CAR decimal ]]

\end{tcolorbox}

\begin{tcolorbox}[
    enhanced,
    breakable,
    colback=orange!10!white, 
    colframe=blue!5!black,
    arc=2mm,
    boxrule=1pt,
    title={\bfseries Audio-Text Models},
    coltitle=white,
    attach boxed title to top left={yshift=-2mm, xshift=3mm},
    boxed title style={enhanced, colback=blue!5!black, colframe=blue!5!black, arc=2mm, boxrule=0pt},
    top=0.5mm, 
    left=1mm, 
    right=1mm, 
    bottom=0.5mm,
]
\setlength{\parskip}{2pt}
\small\vskip8pt

You are an expert quantitative equity analyst. Your task is to predict the stock's Cumulative Abnormal Return (CAR) for a company based on its earnings call transcript and earnings call mp3 audio.

Transcript:

\{text\_transcript\}

Instructions:

1. Predict the Cumulative Abnormal Return (CAR) associated with this earnings event. CAR measures the sum of the differences between the stock's actual return and its expected return over the event window.

2. Provide 1-2 sentences explaining the key driver.

3. Your CAR prediction must be represented as a decimal.

**Answer format**

Rationale = [[ 1-2 sentences rationale ]]

Prediction = [[ CAR decimal ]]

\end{tcolorbox}

\begin{tcolorbox}[
    enhanced,
    breakable,
    colback=orange!10!white, 
    colframe=blue!5!black,
    arc=2mm,
    boxrule=1pt,
    title={\bfseries Any-to-Any Models},
    coltitle=white,
    attach boxed title to top left={yshift=-2mm, xshift=3mm},
    boxed title style={enhanced, colback=blue!5!black, colframe=blue!5!black, arc=2mm, boxrule=0pt},
    top=0.5mm, 
    left=1mm, 
    right=1mm, 
    bottom=0.5mm,
]
\setlength{\parskip}{2pt}
\small\vskip8pt

You are an expert quantitative equity analyst. Your task is to predict the stock's Cumulative Abnormal Return (CAR) for a company based on its earnings call transcript, earnings call audio, and the accompanying presentation slides.

Transcript:

\{text\_transcript\}

Instructions:

1. Predict the Cumulative Abnormal Return (CAR) associated with this earnings event. CAR measures the sum of the differences between the stock's actual return and its expected return over the event window.

2. Provide 1-2 sentences explaining the key driver.

3. Your CAR prediction must be represented as a decimal.

**Answer format**

Rationale = [[ 1-2 sentences rationale ]]

Prediction = [[ CAR decimal ]]

\end{tcolorbox}

\subsection{ret\_eom}

%%%%%%%%%%%%%%%%%%%%%%%%%%

\begin{tcolorbox}[
    enhanced,
    breakable,
    colback=orange!10!white, 
    colframe=blue!5!black,
    arc=2mm,
    boxrule=1pt,
    title={\bfseries Image-Text Models},
    coltitle=white,
    attach boxed title to top left={yshift=-2mm, xshift=3mm},
    boxed title style={enhanced, colback=blue!5!black, colframe=blue!5!black, arc=2mm, boxrule=0pt},
    top=0.5mm, 
    left=1mm, 
    right=1mm, 
    bottom=0.5mm,
]
\setlength{\parskip}{2pt}
\small\vskip8pt

You are an expert quantitative fixed-income analyst. Your task is to predict the End-of-Month corporate bond return (ret\_eom) for a company based on its earnings call transcript and the accompanying presentation slides.

Transcript:

\{text\_transcript\}

Instructions:

1. Predict the ret\_eom (End-of-Month corporate bond return). It represents the percentage gain or loss of an investment or business value over the course of one month.

2. Provide 1-2 sentences explaining the key driver.

3. Your ret\_eom prediction must be represented as a decimal (e.g., 0.052 or -0.014).

**Answer format**

Rationale = [[ 1-2 sentences rationale ]]

Prediction = [[ ret\_eom decimal ]]

\end{tcolorbox}

%%%%%%%%%%%%%%%%%%%%%%%%%%

\begin{tcolorbox}[
    enhanced,
    breakable,
    colback=orange!10!white, 
    colframe=blue!5!black,
    arc=2mm,
    boxrule=1pt,
    title={\bfseries Audio-Text Models},
    coltitle=white,
    attach boxed title to top left={yshift=-2mm, xshift=3mm},
    boxed title style={enhanced, colback=blue!5!black, colframe=blue!5!black, arc=2mm, boxrule=0pt},
    top=0.5mm, 
    left=1mm, 
    right=1mm, 
    bottom=0.5mm,
]
\setlength{\parskip}{2pt}
\small\vskip8pt

You are an expert quantitative fixed-income analyst. Your task is to predict the End-of-Month corporate bond return (ret\_eom) for a company based on its earnings call transcript and earnings call mp3 audio.

Transcript:

\{text\_transcript\}

Instructions:

1. Predict the ret\_eom (End-of-Month corporate bond return). It represents the percentage gain or loss of an investment or business value over the course of one month.

2. Provide 1-2 sentences explaining the key driver.

3. Your ret\_eom prediction must be represented as a decimal (e.g., 0.052 or -0.014).

**Answer format**

Rationale = [[ 1-2 sentences rationale ]]

Prediction = [[ ret\_eom decimal ]]

\end{tcolorbox}

%%%%%%%%%%%%%%%%%%%%%%%%%%

\begin{tcolorbox}[
    enhanced,
    breakable,
    colback=orange!10!white, 
    colframe=blue!5!black,
    arc=2mm,
    boxrule=1pt,
    title={\bfseries Any-to-Any Models},
    coltitle=white,
    attach boxed title to top left={yshift=-2mm, xshift=3mm},
    boxed title style={enhanced, colback=blue!5!black, colframe=blue!5!black, arc=2mm, boxrule=0pt},
    top=0.5mm, 
    left=1mm, 
    right=1mm, 
    bottom=0.5mm,
]
\setlength{\parskip}{2pt}
\small\vskip8pt

You are an expert quantitative fixed-income analyst. Your task is to predict the End-of-Month corporate bond return (ret\_eom) for a company based on its earnings call transcript, earnings call audio, and the accompanying presentation slides.

Transcript:

\{text\_transcript\}

Instructions:

1. Predict the ret\_eom (End-of-Month corporate bond return). It represents the percentage gain or loss of an investment or business value over the course of one month.

2. Provide 1-2 sentences explaining the key driver.

3. Your ret\_eom prediction must be represented as a decimal (e.g., 0.052 or -0.014).

**Answer format**

Rationale = [[ 1-2 sentences rationale ]]

Prediction = [[ ret\_eom decimal ]]

\end{tcolorbox}

\subsection{ret\_l5m}

%%%%%%%%%%%%%%%%%%%%%%%%%%

\begin{tcolorbox}[
    enhanced,
    breakable,
    colback=orange!10!white, 
    colframe=blue!5!black,
    arc=2mm,
    boxrule=1pt,
    title={\bfseries Image-Text Models},
    coltitle=white,
    attach boxed title to top left={yshift=-2mm, xshift=3mm},
    boxed title style={enhanced, colback=blue!5!black, colframe=blue!5!black, arc=2mm, boxrule=0pt},
    top=0.5mm, 
    left=1mm, 
    right=1mm, 
    bottom=0.5mm,
]
\setlength{\parskip}{2pt}
\small\vskip8pt

You are an expert quantitative fixed-income analyst. Your task is to predict the corporate bond return for the last 5 days of the month (ret\_l5m) for a company based on its earnings call transcript and the accompanying presentation slides.

Transcript:

\{text\_transcript\}

Instructions:

1. Predict the ret\_l5m (Return Last 5 Days of Month). It represents the percentage gain or loss of the corporate bond during the final 5 trading days of the month.

2. Provide 1-2 sentences explaining the key driver.

3. Your ret\_l5m prediction must be represented as a decimal (e.g., 0.052 or -0.014).

**Answer format**

Rationale = [[ 1-2 sentences rationale ]]

Prediction = [[ ret\_l5m decimal ]]

\end{tcolorbox}

%%%%%%%%%%%%%%%%%%%%%%%%%%

\begin{tcolorbox}[
    enhanced,
    breakable,
    colback=orange!10!white, 
    colframe=blue!5!black,
    arc=2mm,
    boxrule=1pt,
    title={\bfseries Audio-Text Models},
    coltitle=white,
    attach boxed title to top left={yshift=-2mm, xshift=3mm},
    boxed title style={enhanced, colback=blue!5!black, colframe=blue!5!black, arc=2mm, boxrule=0pt},
    top=0.5mm, 
    left=1mm, 
    right=1mm, 
    bottom=0.5mm,
]
\setlength{\parskip}{2pt}
\small\vskip8pt

You are an expert quantitative fixed-income analyst. Your task is to predict the corporate bond return for the last 5 days of the month (ret\_l5m) for a company based on its earnings call transcript and earnings call mp3 audio.

Transcript:

\{text\_transcript\}

Instructions:

1. Predict the ret\_l5m (Return Last 5 Days of Month). It represents the percentage gain or loss of the corporate bond during the final 5 trading days of the month.

2. Provide 1-2 sentences explaining the key driver.

3. Your ret\_l5m prediction must be represented as a decimal (e.g., 0.052 or -0.014).

**Answer format**

Rationale = [[ 1-2 sentences rationale ]]

Prediction = [[ ret\_l5m decimal ]]

\end{tcolorbox}

%%%%%%%%%%%%%%%%%%%%%%%%%%

\begin{tcolorbox}[
    enhanced,
    breakable,
    colback=orange!10!white, 
    colframe=blue!5!black,
    arc=2mm,
    boxrule=1pt,
    title={\bfseries Any-to-Any Models},
    coltitle=white,
    attach boxed title to top left={yshift=-2mm, xshift=3mm},
    boxed title style={enhanced, colback=blue!5!black, colframe=blue!5!black, arc=2mm, boxrule=0pt},
    top=0.5mm, 
    left=1mm, 
    right=1mm, 
    bottom=0.5mm,
]
\setlength{\parskip}{2pt}
\small\vskip8pt

You are an expert quantitative fixed-income analyst. Your task is to predict the corporate bond return for the last 5 days of the month (ret\_l5m) for a company based on its earnings call transcript, earnings call audio, and the accompanying presentation slides.

Transcript:

\{text\_transcript\}

Instructions:

1. Predict the ret\_l5m (Return Last 5 Days of Month). It represents the percentage gain or loss of the corporate bond during the final 5 trading days of the month.

2. Provide 1-2 sentences explaining the key driver.

3. Your ret\_l5m prediction must be represented as a decimal (e.g., 0.052 or -0.014).

**Answer format**

Rationale = [[ 1-2 sentences rationale ]]

Prediction = [[ ret\_l5m decimal ]]

\end{tcolorbox}

\subsection{ret\_ldm}

%%%%%%%%%%%%%%%%%%%%%%%%%%

\begin{tcolorbox}[
    enhanced,
    breakable,
    colback=orange!10!white, 
    colframe=blue!5!black,
    arc=2mm,
    boxrule=1pt,
    title={\bfseries Image-Text Models},
    coltitle=white,
    attach boxed title to top left={yshift=-2mm, xshift=3mm},
    boxed title style={enhanced, colback=blue!5!black, colframe=blue!5!black, arc=2mm, boxrule=0pt},
    top=0.5mm, 
    left=1mm, 
    right=1mm, 
    bottom=0.5mm,
]
\setlength{\parskip}{2pt}
\small\vskip8pt

You are an expert quantitative fixed-income analyst. Your task is to predict the corporate bond return specifically on the last trading day of the month (ret\_ldm) for a company based on its earnings call transcript and the accompanying presentation slides.

Transcript:

\{text\_transcript\}

Instructions:

1. Predict the ret\_ldm (Return Last Day of Month). It represents the percentage gain or loss of the corporate bond exactly on the final trading day of the month.

2. Provide 1-2 sentences explaining the key driver.

3. Your ret\_ldm prediction must be represented as a decimal (e.g., 0.052 or -0.014).

**Answer format**

Rationale = [[ 1-2 sentences rationale ]]

Prediction = [[ ret\_ldm decimal ]]

\end{tcolorbox}

%%%%%%%%%%%%%%%%%%%%%%%%%%

\begin{tcolorbox}[
    enhanced,
    breakable,
    colback=orange!10!white, 
    colframe=blue!5!black,
    arc=2mm,
    boxrule=1pt,
    title={\bfseries Audio-Text Models},
    coltitle=white,
    attach boxed title to top left={yshift=-2mm, xshift=3mm},
    boxed title style={enhanced, colback=blue!5!black, colframe=blue!5!black, arc=2mm, boxrule=0pt},
    top=0.5mm, 
    left=1mm, 
    right=1mm, 
    bottom=0.5mm,
]
\setlength{\parskip}{2pt}
\small\vskip8pt

You are an expert quantitative fixed-income analyst. Your task is to predict the corporate bond return specifically on the last trading day of the month (ret\_ldm) for a company based on its earnings call transcript and earnings call mp3 audio.

Transcript:

\{text\_transcript\}

Instructions:

1. Predict the ret\_ldm (Return Last Day of Month). It represents the percentage gain or loss of the corporate bond exactly on the final trading day of the month.

2. Provide 1-2 sentences explaining the key driver.

3. Your ret\_ldm prediction must be represented as a decimal (e.g., 0.052 or -0.014).

**Answer format**

Rationale = [[ 1-2 sentences rationale ]]

Prediction = [[ ret\_ldm decimal ]]

\end{tcolorbox}

%%%%%%%%%%%%%%%%%%%%%%%%%%

\begin{tcolorbox}[
    enhanced,
    breakable,
    colback=orange!10!white, 
    colframe=blue!5!black,
    arc=2mm,
    boxrule=1pt,
    title={\bfseries Any-to-Any Models},
    coltitle=white,
    attach boxed title to top left={yshift=-2mm, xshift=3mm},
    boxed title style={enhanced, colback=blue!5!black, colframe=blue!5!black, arc=2mm, boxrule=0pt},
    top=0.5mm, 
    left=1mm, 
    right=1mm, 
    bottom=0.5mm,
]
\setlength{\parskip}{2pt}
\small\vskip8pt

You are an expert quantitative fixed-income analyst. Your task is to predict the corporate bond return specifically on the last trading day of the month (ret\_ldm) for a company based on its earnings call transcript, earnings call audio, and the accompanying presentation slides.

Transcript:

\{text\_transcript\}

Instructions:

1. Predict the ret\_ldm (Return Last Day of Month). It represents the percentage gain or loss of the corporate bond exactly on the final trading day of the month.

2. Provide 1-2 sentences explaining the key driver.

3. Your ret\_ldm prediction must be represented as a decimal (e.g., 0.052 or -0.014).

**Answer format**

Rationale = [[ 1-2 sentences rationale ]]

Prediction = [[ ret\_ldm decimal ]]

\end{tcolorbox}

\subsection{vol\_fut\_3d}

%%%%%%%%%%%%%%%%%%%%%%%%%%

\begin{tcolorbox}[
    enhanced,
    breakable,
    colback=orange!10!white, 
    colframe=blue!5!black,
    arc=2mm,
    boxrule=1pt,
    title={\bfseries Image-Text Models},
    coltitle=white,
    attach boxed title to top left={yshift=-2mm, xshift=3mm},
    boxed title style={enhanced, colback=blue!5!black, colframe=blue!5!black, arc=2mm, boxrule=0pt},
    top=0.5mm, 
    left=1mm, 
    right=1mm, 
    bottom=0.5mm,
]
\setlength{\parskip}{2pt}
\small\vskip8pt

You are an expert quantitative analyst. Your task is to predict the post-event 3-day volatility (vol\_fut\_3d) for a company based on its earnings call transcript and the accompanying presentation slides.

Transcript:

\{text\_transcript\}

Instructions:

1. Predict the vol\_fut\_3d (Volatility Future 3 Days). This represents the expected price turbulence, uncertainty, and fluctuation of the asset over the 3 trading days immediately following this earnings call.

2. Provide 1-2 sentences explaining the key driver.

3. Your vol\_fut\_3d estimate must be represented as a decimal.

**Answer format**

Rationale = [[ 1-2 sentences rationale ]]

Prediction = [[ vol\_fut\_3d decimal ]]

\end{tcolorbox}

%%%%%%%%%%%%%%%%%%%%%%%%%%

\begin{tcolorbox}[
    enhanced,
    breakable,
    colback=orange!10!white, 
    colframe=blue!5!black,
    arc=2mm,
    boxrule=1pt,
    title={\bfseries Audio-Text Models},
    coltitle=white,
    attach boxed title to top left={yshift=-2mm, xshift=3mm},
    boxed title style={enhanced, colback=blue!5!black, colframe=blue!5!black, arc=2mm, boxrule=0pt},
    top=0.5mm, 
    left=1mm, 
    right=1mm, 
    bottom=0.5mm,
]
\setlength{\parskip}{2pt}
\small\vskip8pt

You are an expert quantitative analyst. Your task is to predict the post-event 3-day volatility (vol\_fut\_3d) for a company based on its earnings call transcript and earnings call mp3 audio.

Transcript:

\{text\_transcript\}

Instructions:

1. Predict the vol\_fut\_3d (Volatility Future 3 Days). This represents the expected price turbulence, uncertainty, and fluctuation of the asset over the 3 trading days immediately following this earnings call.

2. Provide 1-2 sentences explaining the key driver.

3. Your vol\_fut\_3d estimate must be represented as a decimal.

**Answer format**

Rationale = [[ 1-2 sentences rationale ]]

Prediction = [[ vol\_fut\_3d decimal ]]

\end{tcolorbox}

%%%%%%%%%%%%%%%%%%%%%%%%%%

\begin{tcolorbox}[
    enhanced,
    breakable,
    colback=orange!10!white, 
    colframe=blue!5!black,
    arc=2mm,
    boxrule=1pt,
    title={\bfseries Any-to-Any Models},
    coltitle=white,
    attach boxed title to top left={yshift=-2mm, xshift=3mm},
    boxed title style={enhanced, colback=blue!5!black, colframe=blue!5!black, arc=2mm, boxrule=0pt},
    top=0.5mm, 
    left=1mm, 
    right=1mm, 
    bottom=0.5mm,
]
\setlength{\parskip}{2pt}
\small\vskip8pt

You are an expert quantitative analyst. Your task is to predict the post-event 3-day volatility (vol\_fut\_3d) for a company based on its earnings call transcript, earnings call audio, and the accompanying presentation slides.

Transcript:

\{text\_transcript\}

Instructions:

1. Predict the vol\_fut\_3d (Volatility Future 3 Days). This represents the expected price turbulence, uncertainty, and fluctuation of the asset over the 3 trading days immediately following this earnings call.

2. Provide 1-2 sentences explaining the key driver.

3. Your vol\_fut\_3d estimate must be represented as a decimal.

**Answer format**

Rationale = [[ 1-2 sentences rationale ]]

Prediction = [[ vol\_fut\_3d decimal ]]

\end{tcolorbox}

\subsection{vol\_fut\_7d}

%%%%%%%%%%%%%%%%%%%%%%%%%%

\begin{tcolorbox}[
    enhanced,
    breakable,
    colback=orange!10!white, 
    colframe=blue!5!black,
    arc=2mm,
    boxrule=1pt,
    title={\bfseries Image-Text Models},
    coltitle=white,
    attach boxed title to top left={yshift=-2mm, xshift=3mm},
    boxed title style={enhanced, colback=blue!5!black, colframe=blue!5!black, arc=2mm, boxrule=0pt},
    top=0.5mm, 
    left=1mm, 
    right=1mm, 
    bottom=0.5mm,
]
\setlength{\parskip}{2pt}
\small\vskip8pt

You are an expert quantitative analyst. Your task is to predict the post-event 7-day volatility (vol\_fut\_7d) for a company based on its earnings call transcript and the accompanying presentation slides.

Transcript:

\{text\_transcript\}

Instructions:

1. Predict the vol\_fut\_7d (Volatility Future 7 Days). This represents the expected price turbulence, uncertainty, and fluctuation of the asset over the 7 trading days immediately following this earnings call.

2. Provide 1-2 sentences explaining the key driver.

3. Your vol\_fut\_7d estimate must be represented as a decimal.

**Answer format**

Rationale = [[ 1-2 sentences rationale ]]

Prediction = [[ vol\_fut\_7d decimal ]]

\end{tcolorbox}

%%%%%%%%%%%%%%%%%%%%%%%%%%

\begin{tcolorbox}[
    enhanced,
    breakable,
    colback=orange!10!white, 
    colframe=blue!5!black,
    arc=2mm,
    boxrule=1pt,
    title={\bfseries Audio-Text Models},
    coltitle=white,
    attach boxed title to top left={yshift=-2mm, xshift=3mm},
    boxed title style={enhanced, colback=blue!5!black, colframe=blue!5!black, arc=2mm, boxrule=0pt},
    top=0.5mm, 
    left=1mm, 
    right=1mm, 
    bottom=0.5mm,
]
\setlength{\parskip}{2pt}
\small\vskip8pt

You are an expert quantitative analyst. Your task is to predict the post-event 7-day volatility (vol\_fut\_7d) for a company based on its earnings call transcript and earnings call mp3 audio.

Transcript:

\{text\_transcript\}

Instructions:

1. Predict the vol\_fut\_7d (Volatility Future 7 Days). This represents the expected price turbulence, uncertainty, and fluctuation of the asset over the 7 trading days immediately following this earnings call.

2. Provide 1-2 sentences explaining the key driver.

3. Your vol\_fut\_7d estimate must be represented as a decimal.

**Answer format**

Rationale = [[ 1-2 sentences rationale ]]

Prediction = [[ vol\_fut\_7d decimal ]]

\end{tcolorbox}

%%%%%%%%%%%%%%%%%%%%%%%%%%

\begin{tcolorbox}[
    enhanced,
    breakable,
    colback=orange!10!white, 
    colframe=blue!5!black,
    arc=2mm,
    boxrule=1pt,
    title={\bfseries Any-to-Any Models},
    coltitle=white,
    attach boxed title to top left={yshift=-2mm, xshift=3mm},
    boxed title style={enhanced, colback=blue!5!black, colframe=blue!5!black, arc=2mm, boxrule=0pt},
    top=0.5mm, 
    left=1mm, 
    right=1mm, 
    bottom=0.5mm,
]
\setlength{\parskip}{2pt}
\small\vskip8pt

You are an expert quantitative analyst. Your task is to predict the post-event 7-day volatility (vol\_fut\_7d) for a company based on its earnings call transcript, earnings call audio, and the accompanying presentation slides.

Transcript:

\{text\_transcript\}

Instructions:

1. Predict the vol\_fut\_7d (Volatility Future 7 Days). This represents the expected price turbulence, uncertainty, and fluctuation of the asset over the 7 trading days immediately following this earnings call.

2. Provide 1-2 sentences explaining the key driver.

3. Your vol\_fut\_7d estimate must be represented as a decimal.

**Answer format**

Rationale = [[ 1-2 sentences rationale ]]

Prediction = [[ vol\_fut\_7d decimal ]]

\end{tcolorbox}

\subsection{vol\_fut\_15d}

%%%%%%%%%%%%%%%%%%%%%%%%%%

\begin{tcolorbox}[
    enhanced,
    breakable,
    colback=orange!10!white, 
    colframe=blue!5!black,
    arc=2mm,
    boxrule=1pt,
    title={\bfseries Image-Text Models},
    coltitle=white,
    attach boxed title to top left={yshift=-2mm, xshift=3mm},
    boxed title style={enhanced, colback=blue!5!black, colframe=blue!5!black, arc=2mm, boxrule=0pt},
    top=0.5mm, 
    left=1mm, 
    right=1mm, 
    bottom=0.5mm,
]
\setlength{\parskip}{2pt}
\small\vskip8pt

You are an expert quantitative analyst. Your task is to predict the post-event 15-day volatility (vol\_fut\_15d) for a company based on its earnings call transcript and the accompanying presentation slides.

Transcript:

\{text\_transcript\}

Instructions:

1. Predict the vol\_fut\_15d (Volatility Future 15 Days). This represents the expected price turbulence, uncertainty, and fluctuation of the asset over the 15 trading days immediately following this earnings call.

2. Provide 1-2 sentences explaining the key driver.

3. Your vol\_fut\_15d estimate must be represented as a decimal.

**Answer format**

Rationale = [[ 1-2 sentences rationale ]]

Prediction = [[ vol\_fut\_15d decimal ]]

\end{tcolorbox}

%%%%%%%%%%%%%%%%%%%%%%%%%%

\begin{tcolorbox}[
    enhanced,
    breakable,
    colback=orange!10!white, 
    colframe=blue!5!black,
    arc=2mm,
    boxrule=1pt,
    title={\bfseries Audio-Text Models},
    coltitle=white,
    attach boxed title to top left={yshift=-2mm, xshift=3mm},
    boxed title style={enhanced, colback=blue!5!black, colframe=blue!5!black, arc=2mm, boxrule=0pt},
    top=0.5mm, 
    left=1mm, 
    right=1mm, 
    bottom=0.5mm,
]
\setlength{\parskip}{2pt}
\small\vskip8pt

You are an expert quantitative analyst. Your task is to predict the post-event 15-day volatility (vol\_fut\_15d) for a company based on its earnings call transcript and earnings call mp3 audio.

Transcript:

\{text\_transcript\}

Instructions:

1. Predict the vol\_fut\_15d (Volatility Future 15 Days). This represents the expected price turbulence, uncertainty, and fluctuation of the asset over the 15 trading days immediately following this earnings call.

2. Provide 1-2 sentences explaining the key driver.

3. Your vol\_fut\_15d estimate must be represented as a decimal.

**Answer format**

Rationale = [[ 1-2 sentences rationale ]]

Prediction = [[ vol\_fut\_15d decimal ]]

\end{tcolorbox}

%%%%%%%%%%%%%%%%%%%%%%%%%%

\begin{tcolorbox}[
    enhanced,
    breakable,
    colback=orange!10!white, 
    colframe=blue!5!black,
    arc=2mm,
    boxrule=1pt,
    title={\bfseries Any-to-Any Models},
    coltitle=white,
    attach boxed title to top left={yshift=-2mm, xshift=3mm},
    boxed title style={enhanced, colback=blue!5!black, colframe=blue!5!black, arc=2mm, boxrule=0pt},
    top=0.5mm, 
    left=1mm, 
    right=1mm, 
    bottom=0.5mm,
]
\setlength{\parskip}{2pt}
\small\vskip8pt

You are an expert quantitative analyst. Your task is to predict the post-event 15-day volatility (vol\_fut\_15d) for a company based on its earnings call transcript, earnings call audio, and the accompanying presentation slides.

Transcript:

\{text\_transcript\}

Instructions:

1. Predict the vol\_fut\_15d (Volatility Future 15 Days). This represents the expected price turbulence, uncertainty, and fluctuation of the asset over the 15 trading days immediately following this earnings call.

2. Provide 1-2 sentences explaining the key driver.

3. Your vol\_fut\_15d estimate must be represented as a decimal.

**Answer format**

Rationale = [[ 1-2 sentences rationale ]]

Prediction = [[ vol\_fut\_15d decimal ]]

\end{tcolorbox}

\subsection{vol\_fut\_30d}

%%%%%%%%%%%%%%%%%%%%%%%%%%

\begin{tcolorbox}[
    enhanced,
    breakable,
    colback=orange!10!white, 
    colframe=blue!5!black,
    arc=2mm,
    boxrule=1pt,
    title={\bfseries Image-Text Models},
    coltitle=white,
    attach boxed title to top left={yshift=-2mm, xshift=3mm},
    boxed title style={enhanced, colback=blue!5!black, colframe=blue!5!black, arc=2mm, boxrule=0pt},
    top=0.5mm, 
    left=1mm, 
    right=1mm, 
    bottom=0.5mm,
]
\setlength{\parskip}{2pt}
\small\vskip8pt

You are an expert quantitative analyst. Your task is to predict the post-event 30-day volatility (vol\_fut\_30d) for a company based on its earnings call transcript and the accompanying presentation slides.

Transcript:

\{text\_transcript\}

Instructions:

1. Predict the vol\_fut\_30d (Volatility Future 30 Days). This represents the expected price turbulence, uncertainty, and fluctuation of the asset over the 30 trading days immediately following this earnings call.

2. Provide 1-2 sentences explaining the key driver.

3. Your vol\_fut\_30d estimate must be represented as a decimal.

**Answer format**

Rationale = [[ 1-2 sentences rationale ]]

Prediction = [[ vol\_fut\_30d decimal ]]

\end{tcolorbox}

%%%%%%%%%%%%%%%%%%%%%%%%%%

\begin{tcolorbox}[
    enhanced,
    breakable,
    colback=orange!10!white, 
    colframe=blue!5!black,
    arc=2mm,
    boxrule=1pt,
    title={\bfseries Audio-Text Models},
    coltitle=white,
    attach boxed title to top left={yshift=-2mm, xshift=3mm},
    boxed title style={enhanced, colback=blue!5!black, colframe=blue!5!black, arc=2mm, boxrule=0pt},
    top=0.5mm, 
    left=1mm, 
    right=1mm, 
    bottom=0.5mm,
]
\setlength{\parskip}{2pt}
\small\vskip8pt

You are an expert quantitative analyst. Your task is to predict the post-event 30-day volatility (vol\_fut\_30d) for a company based on its earnings call transcript and earnings call mp3 audio.

Transcript:

\{text\_transcript\}

Instructions:

1. Predict the vol\_fut\_30d (Volatility Future 30 Days). This represents the expected price turbulence, uncertainty, and fluctuation of the asset over the 30 trading days immediately following this earnings call.

2. Provide 1-2 sentences explaining the key driver.

3. Your vol\_fut\_30d estimate must be represented as a decimal.

**Answer format**

Rationale = [[ 1-2 sentences rationale ]]

Prediction = [[ vol\_fut\_30d decimal ]]

\end{tcolorbox}

%%%%%%%%%%%%%%%%%%%%%%%%%%

\begin{tcolorbox}[
    enhanced,
    breakable,
    colback=orange!10!white, 
    colframe=blue!5!black,
    arc=2mm,
    boxrule=1pt,
    title={\bfseries Any-to-Any Models},
    coltitle=white,
    attach boxed title to top left={yshift=-2mm, xshift=3mm},
    boxed title style={enhanced, colback=blue!5!black, colframe=blue!5!black, arc=2mm, boxrule=0pt},
    top=0.5mm, 
    left=1mm, 
    right=1mm, 
    bottom=0.5mm,
]
\setlength{\parskip}{2pt}
\small\vskip8pt

You are an expert quantitative analyst. Your task is to predict the post-event 30-day volatility (vol\_fut\_30d) for a company based on its earnings call transcript, earnings call audio, and the accompanying presentation slides.

Transcript:

\{text\_transcript\}

Instructions:

1. Predict the vol\_fut\_30d (Volatility Future 30 Days). This represents the expected price turbulence, uncertainty, and fluctuation of the asset over the 30 trading days immediately following this earnings call.

2. Provide 1-2 sentences explaining the key driver.

3. Your vol\_fut\_30d estimate must be represented as a decimal.

**Answer format**

Rationale = [[ 1-2 sentences rationale ]]

Prediction = [[ vol\_fut\_30d decimal ]]

\end{tcolorbox}

\subsection{vol\_past\_3d}

%%%%%%%%%%%%%%%%%%%%%%%%%%

\begin{tcolorbox}[
    enhanced,
    breakable,
    colback=orange!10!white, 
    colframe=blue!5!black,
    arc=2mm,
    boxrule=1pt,
    title={\bfseries Image-Text Models},
    coltitle=white,
    attach boxed title to top left={yshift=-2mm, xshift=3mm},
    boxed title style={enhanced, colback=blue!5!black, colframe=blue!5!black, arc=2mm, boxrule=0pt},
    top=0.5mm, 
    left=1mm, 
    right=1mm, 
    bottom=0.5mm,
]
\setlength{\parskip}{2pt}
\small\vskip8pt

You are an expert quantitative analyst. Your task is to estimate the pre-event 3-day volatility (vol\_past\_3d) for a company based on its earnings call transcript and the accompanying presentation slides.

Transcript:

\{text\_transcript\}

Instructions:

1. Estimate the vol\_past\_3d (Volatility Past 3 Days). This represents the realized volatility (price turbulence and uncertainty) of the asset over the 3 trading days immediately preceding this earnings call. 

2. Provide 1-2 sentences explaining the key driver.

3. Your vol\_past\_3d estimate must be represented as a decimal.

**Answer format**

Rationale = [[ 1-2 sentences rationale ]]

Prediction = [[ vol\_past\_3d decimal ]]

\end{tcolorbox}

%%%%%%%%%%%%%%%%%%%%%%%%%%

\begin{tcolorbox}[
    enhanced,
    breakable,
    colback=orange!10!white, 
    colframe=blue!5!black,
    arc=2mm,
    boxrule=1pt,
    title={\bfseries Audio-Text Models},
    coltitle=white,
    attach boxed title to top left={yshift=-2mm, xshift=3mm},
    boxed title style={enhanced, colback=blue!5!black, colframe=blue!5!black, arc=2mm, boxrule=0pt},
    top=0.5mm, 
    left=1mm, 
    right=1mm, 
    bottom=0.5mm,
]
\setlength{\parskip}{2pt}
\small\vskip8pt

You are an expert quantitative analyst. Your task is to estimate the pre-event 3-day  volatility (vol\_past\_3d) for a company based on its earnings call transcript and earnings call mp3 audio.

Transcript:

\{text\_transcript\}

Instructions:

1. Estimate the vol\_past\_3d (Volatility Past 3 Days). This represents the realized volatility (price turbulence and uncertainty) of the asset over the 3 trading days immediately preceding this earnings call. 

2. Provide 1-2 sentences explaining the key driver.

3. Your vol\_past\_3d estimate must be represented as a decimal.

**Answer format**

Rationale = [[ 1-2 sentences rationale ]]

Prediction = [[ vol\_past\_3d decimal ]]

\end{tcolorbox}

%%%%%%%%%%%%%%%%%%%%%%%%%%

\begin{tcolorbox}[
    enhanced,
    breakable,
    colback=orange!10!white, 
    colframe=blue!5!black,
    arc=2mm,
    boxrule=1pt,
    title={\bfseries Any-to-Any Models},
    coltitle=white,
    attach boxed title to top left={yshift=-2mm, xshift=3mm},
    boxed title style={enhanced, colback=blue!5!black, colframe=blue!5!black, arc=2mm, boxrule=0pt},
    top=0.5mm, 
    left=1mm, 
    right=1mm, 
    bottom=0.5mm,
]
\setlength{\parskip}{2pt}
\small\vskip8pt

You are an expert quantitative analyst. Your task is to estimate the pre-event 3-day volatility (vol\_past\_3d) for a company based on its earnings call transcript, earnings call audio, and the accompanying presentation slides.

Transcript:

\{text\_transcript\}

Instructions:

1. Estimate the vol\_past\_3d (Volatility Past 3 Days). This represents the realized volatility (price turbulence and uncertainty) of the asset over the 3 trading days immediately preceding this earnings call. 

2. Provide 1-2 sentences explaining the key driver.

3. Your vol\_past\_3d estimate must be represented as a decimal.

**Answer format**

Rationale = [[ 1-2 sentences rationale ]]

Prediction = [[ vol\_past\_3d decimal ]]

\end{tcolorbox}

\subsection{vol\_past\_7d}

%%%%%%%%%%%%%%%%%%%%%%%%%%

\begin{tcolorbox}[
    enhanced,
    breakable,
    colback=orange!10!white, 
    colframe=blue!5!black,
    arc=2mm,
    boxrule=1pt,
    title={\bfseries Image-Text Models},
    coltitle=white,
    attach boxed title to top left={yshift=-2mm, xshift=3mm},
    boxed title style={enhanced, colback=blue!5!black, colframe=blue!5!black, arc=2mm, boxrule=0pt},
    top=0.5mm, 
    left=1mm, 
    right=1mm, 
    bottom=0.5mm,
]
\setlength{\parskip}{2pt}
\small\vskip8pt

You are an expert quantitative analyst. Your task is to estimate the pre-event 7-day volatility (vol\_past\_7d) for a company based on its earnings call transcript and the accompanying presentation slides.

Transcript:

\{text\_transcript\}

Instructions:

1. Estimate the vol\_past\_7d (Volatility Past 7 Days). This represents the realized volatility (price turbulence and uncertainty) of the asset over the 7 trading days immediately preceding this earnings call. 

2. Provide 1-2 sentences explaining the key driver.

3. Your vol\_past\_7d estimate must be represented as a decimal.

**Answer format**

Rationale = [[ 1-2 sentences rationale ]]

Prediction = [[ vol\_past\_7d decimal ]]

\end{tcolorbox}

%%%%%%%%%%%%%%%%%%%%%%%%%%

\begin{tcolorbox}[
    enhanced,
    breakable,
    colback=orange!10!white, 
    colframe=blue!5!black,
    arc=2mm,
    boxrule=1pt,
    title={\bfseries Audio-Text Models},
    coltitle=white,
    attach boxed title to top left={yshift=-2mm, xshift=3mm},
    boxed title style={enhanced, colback=blue!5!black, colframe=blue!5!black, arc=2mm, boxrule=0pt},
    top=0.5mm, 
    left=1mm, 
    right=1mm, 
    bottom=0.5mm,
]
\setlength{\parskip}{2pt}
\small\vskip8pt

You are an expert quantitative analyst. Your task is to estimate the pre-event 7-day volatility (vol\_past\_7d) for a company based on its earnings call transcript and earnings call mp3 audio.

Transcript:

\{text\_transcript\}

Instructions:

1. Estimate the vol\_past\_7d (Volatility Past 7 Days). This represents the realized volatility (price turbulence and uncertainty) of the asset over the 7 trading days immediately preceding this earnings call. 

2. Provide 1-2 sentences explaining the key driver.

3. Your vol\_past\_7d estimate must be represented as a decimal.

**Answer format**

Rationale = [[ 1-2 sentences rationale ]]

Prediction = [[ vol\_past\_7d decimal ]]

\end{tcolorbox}

%%%%%%%%%%%%%%%%%%%%%%%%%%

\begin{tcolorbox}[
    enhanced,
    breakable,
    colback=orange!10!white, 
    colframe=blue!5!black,
    arc=2mm,
    boxrule=1pt,
    title={\bfseries Any-to-Any Models},
    coltitle=white,
    attach boxed title to top left={yshift=-2mm, xshift=3mm},
    boxed title style={enhanced, colback=blue!5!black, colframe=blue!5!black, arc=2mm, boxrule=0pt},
    top=0.5mm, 
    left=1mm, 
    right=1mm, 
    bottom=0.5mm,
]
\setlength{\parskip}{2pt}
\small\vskip8pt

You are an expert quantitative analyst. Your task is to estimate the pre-event 7-day volatility (vol\_past\_7d) for a company based on its earnings call transcript, earnings call audio, and the accompanying presentation slides.

Transcript:

\{text\_transcript\}

Instructions:

1. Estimate the vol\_past\_7d (Volatility Past 7 Days). This represents the realized volatility (price turbulence and uncertainty) of the asset over the 7 trading days immediately preceding this earnings call. 

2. Provide 1-2 sentences explaining the key driver.

3. Your vol\_past\_7d estimate must be represented as a decimal.

**Answer format**

Rationale = [[ 1-2 sentences rationale ]]

Prediction = [[ vol\_past\_7d decimal ]]

\end{tcolorbox}

\subsection{vol\_past\_15d}

%%%%%%%%%%%%%%%%%%%%%%%%%%

\begin{tcolorbox}[
    enhanced,
    breakable,
    colback=orange!10!white, 
    colframe=blue!5!black,
    arc=2mm,
    boxrule=1pt,
    title={\bfseries Image-Text Models},
    coltitle=white,
    attach boxed title to top left={yshift=-2mm, xshift=3mm},
    boxed title style={enhanced, colback=blue!5!black, colframe=blue!5!black, arc=2mm, boxrule=0pt},
    top=0.5mm, 
    left=1mm, 
    right=1mm, 
    bottom=0.5mm,
]
\setlength{\parskip}{2pt}
\small\vskip8pt

You are an expert quantitative analyst. Your task is to estimate the pre-event 15-day volatility (vol\_past\_15d) for a company based on its earnings call transcript and the accompanying presentation slides.

Transcript:

\{text\_transcript\}

Instructions:

1. Estimate the vol\_past\_15d (Volatility Past 15 Days). This represents the realized volatility (price turbulence and uncertainty) of the asset over the 15 trading days immediately preceding this earnings call. 

2. Provide 1-2 sentences explaining the key driver.

3. Your vol\_past\_15d estimate must be represented as a decimal.

**Answer format**

Rationale = [[ 1-2 sentences rationale ]]

Prediction = [[ vol\_past\_15d decimal ]]

\end{tcolorbox}

%%%%%%%%%%%%%%%%%%%%%%%%%%

\begin{tcolorbox}[
    enhanced,
    breakable,
    colback=orange!10!white, 
    colframe=blue!5!black,
    arc=2mm,
    boxrule=1pt,
    title={\bfseries Audio-Text Models},
    coltitle=white,
    attach boxed title to top left={yshift=-2mm, xshift=3mm},
    boxed title style={enhanced, colback=blue!5!black, colframe=blue!5!black, arc=2mm, boxrule=0pt},
    top=0.5mm, 
    left=1mm, 
    right=1mm, 
    bottom=0.5mm,
]
\setlength{\parskip}{2pt}
\small\vskip8pt

You are an expert quantitative analyst. Your task is to estimate the pre-event 15-day volatility (vol\_past\_15d) for a company based on its earnings call transcript and earnings call mp3 audio.

Transcript:

\{text\_transcript\}

Instructions:

1. Estimate the vol\_past\_15d (Volatility Past 15 Days). This represents the realized volatility (price turbulence and uncertainty) of the asset over the 15 trading days immediately preceding this earnings call. 

2. Provide 1-2 sentences explaining the key driver.

3. Your vol\_past\_15d estimate must be represented as a decimal.

**Answer format**

Rationale = [[ 1-2 sentences rationale ]]

Prediction = [[ vol\_past\_15d decimal ]]

\end{tcolorbox}

%%%%%%%%%%%%%%%%%%%%%%%%%%

\begin{tcolorbox}[
    enhanced,
    breakable,
    colback=orange!10!white, 
    colframe=blue!5!black,
    arc=2mm,
    boxrule=1pt,
    title={\bfseries Any-to-Any Models},
    coltitle=white,
    attach boxed title to top left={yshift=-2mm, xshift=3mm},
    boxed title style={enhanced, colback=blue!5!black, colframe=blue!5!black, arc=2mm, boxrule=0pt},
    top=0.5mm, 
    left=1mm, 
    right=1mm, 
    bottom=0.5mm,
]
\setlength{\parskip}{2pt}
\small\vskip8pt

You are an expert quantitative analyst. Your task is to estimate the pre-event 15-day volatility (vol\_past\_15d) for a company based on its earnings call transcript, earnings call audio, and the accompanying presentation slides.

Transcript:

\{text\_transcript\}

Instructions:

1. Estimate the vol\_past\_15d (Volatility Past 15 Days). This represents the realized volatility (price turbulence and uncertainty) of the asset over the 15 trading days immediately preceding this earnings call. 

2. Provide 1-2 sentences explaining the key driver.

3. Your vol\_past\_15d estimate must be represented as a decimal.

**Answer format**

Rationale = [[ 1-2 sentences rationale ]]

Prediction = [[ vol\_past\_15d decimal ]]

\end{tcolorbox}

\subsection{vol\_past\_30d}

%%%%%%%%%%%%%%%%%%%%%%%%%%

\begin{tcolorbox}[
    enhanced,
    breakable,
    colback=orange!10!white, 
    colframe=blue!5!black,
    arc=2mm,
    boxrule=1pt,
    title={\bfseries Image-Text Models},
    coltitle=white,
    attach boxed title to top left={yshift=-2mm, xshift=3mm},
    boxed title style={enhanced, colback=blue!5!black, colframe=blue!5!black, arc=2mm, boxrule=0pt},
    top=0.5mm, 
    left=1mm, 
    right=1mm, 
    bottom=0.5mm,
]
\setlength{\parskip}{2pt}
\small\vskip8pt

You are an expert quantitative analyst. Your task is to estimate the pre-event 30-day volatility (vol\_past\_30d) for a company based on its earnings call transcript and the accompanying presentation slides.

Transcript:

\{text\_transcript\}

Instructions:

1. Estimate the vol\_past\_30d (Volatility Past 30 Days). This represents the realized volatility (price turbulence and uncertainty) of the asset over the 30 trading days immediately preceding this earnings call. 

2. Provide 1-2 sentences explaining the key driver.

3. Your vol\_past\_30d estimate must be represented as a decimal.

**Answer format**

Rationale = [[ 1-2 sentences rationale ]]

Prediction = [[ vol\_past\_30d decimal ]]

\end{tcolorbox}

%%%%%%%%%%%%%%%%%%%%%%%%%%

\begin{tcolorbox}[
    enhanced,
    breakable,
    colback=orange!10!white, 
    colframe=blue!5!black,
    arc=2mm,
    boxrule=1pt,
    title={\bfseries Audio-Text Models},
    coltitle=white,
    attach boxed title to top left={yshift=-2mm, xshift=3mm},
    boxed title style={enhanced, colback=blue!5!black, colframe=blue!5!black, arc=2mm, boxrule=0pt},
    top=0.5mm, 
    left=1mm, 
    right=1mm, 
    bottom=0.5mm,
]
\setlength{\parskip}{2pt}
\small\vskip8pt

You are an expert quantitative analyst. Your task is to estimate the pre-event 30-day volatility (vol\_past\_30d) for a company based on its earnings call transcript and earnings call mp3 audio.

Transcript:

\{text\_transcript\}

Instructions:

1. Estimate the vol\_past\_30d (Volatility Past 30 Days). This represents the realized volatility (price turbulence and uncertainty) of the asset over the 30 trading days immediately preceding this earnings call. 

2. Provide 1-2 sentences explaining the key driver.

3. Your vol\_past\_30d estimate must be represented as a decimal.

**Answer format**

Rationale = [[ 1-2 sentences rationale ]]

Prediction = [[ vol\_past\_30d decimal ]]

\end{tcolorbox}

%%%%%%%%%%%%%%%%%%%%%%%%%%

\begin{tcolorbox}[
    enhanced,
    breakable,
    colback=orange!10!white, 
    colframe=blue!5!black,
    arc=2mm,
    boxrule=1pt,
    title={\bfseries Any-to-Any Models},
    coltitle=white,
    attach boxed title to top left={yshift=-2mm, xshift=3mm},
    boxed title style={enhanced, colback=blue!5!black, colframe=blue!5!black, arc=2mm, boxrule=0pt},
    top=0.5mm, 
    left=1mm, 
    right=1mm, 
    bottom=0.5mm,
]
\setlength{\parskip}{2pt}
\small\vskip8pt

You are an expert quantitative analyst. Your task is to estimate the pre-event 30-day volatility (vol\_past\_30d) for a company based on its earnings call transcript, earnings call audio, and the accompanying presentation slides.

Transcript:

\{text\_transcript\}

Instructions:

1. Estimate the vol\_past\_30d (Volatility Past 30 Days). This represents the realized volatility (price turbulence and uncertainty) of the asset over the 30 trading days immediately preceding this earnings call. 

2. Provide 1-2 sentences explaining the key driver.

3. Your vol\_past\_30d estimate must be represented as a decimal.

**Answer format**

Rationale = [[ 1-2 sentences rationale ]]

Prediction = [[ vol\_past\_30d decimal ]]

\end{tcolorbox}

\begin{figure*}[ht]
    \centering
    \includegraphics[width=1\linewidth]{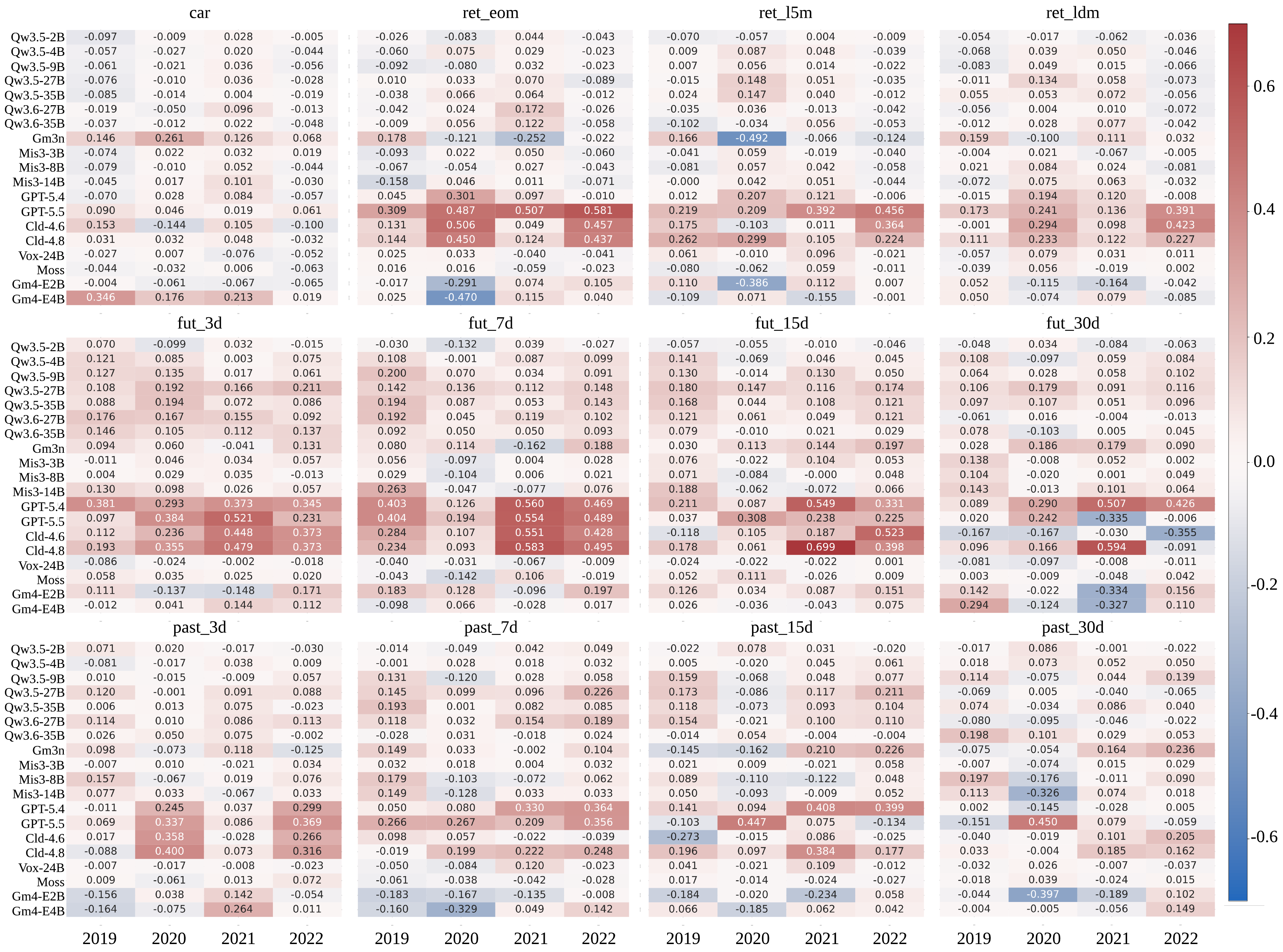}
    \caption{Annual Pearson Correlation Trends (2019–2022) on the MM-FinEval Benchmark. The heatmap illustrates how models' ability to capture linear relationships shifts over time.}
    \label{fig:pearson_trend}
\end{figure*}

\section{Pearson Correlation Temporal Observations}
\label{app:pearson_temporal}
We evaluated the Pearson correlation coefficient across the 2019–2022 timeline. Figure \ref{fig:pearson_trend} indicates that proprietary models improve their ability to capture linear relationships over time. For example, on the ret\_eom task, GPT-5.5 exhibits a strict upward trajectory, scaling from a correlation of 0.309 in 2019 to 0.581 by 2022. Similarly, Claude 4.8 improves from 0.144 in 2019 to 0.437 in 2022 on the exact same task. This temporal strengthening suggests that massive, closed-source models are highly adept at mapping evolving macroeconomic environments to relative asset rankings. In contrast, open-source models demonstrate no such temporal adaptation. Models like Qwen 3.5 and Mistral 3 remain near-zero correlations on ret\_eom across the four-year span.

% Finally, examining the volatility tasks reveals a clear temporal degradation as the prediction horizon expands. This is illustrated by GPT and Claude's performance across the post-event volatility windows: these models achieve a strong correlation for 3 and 7 day post-event volatility (fut\_3d and fut\_7d), which steadily drops at 15 and 30 days (fut\_15d and fut\_30d). This progressive decay aligns with our previous findings in Section \ref{sec:pearson_correlation} and the information half-life theory that the predictive signal contained within a corporate disclosure is rapidly diluted by unpredictable market factors over longer temporal windows.

\begin{table*}[ht]
\centering
\renewcommand{\arraystretch}{1.4}
\resizebox{\textwidth}{!}{
\begin{tabular}{ll ccccccccccccccc cc cc}
\toprule
\toprule
\multirow{2}{*}{Label} & \multirow{2}{*}{Year} & \multicolumn{15}{c}{\textit{Image-Text}} & \multicolumn{2}{c}{\textit{Audio-Text}} & \multicolumn{2}{c}{\textit{Any-to-Any}} \\ \cmidrule(lr){3-17} \cmidrule(lr){18-19} \cmidrule(lr){20-21}
 & & Qw3.5-2B & Qw3.5-4B & Qw3.5-9B & Qw3.5-27B & Qw3.5-35B & Qw3.6-27B & Qw3.6-35B & Gm3n & Mis3-3B & Mis3-8B & Mis3-14B & GPT-5.4 & GPT-5.5 & Cld-4.6 & Cld-4.8 & Vox-24B & Moss & Gm4-E2B & Gm4-E4B \\ \hline
\multirow{4}{*}{car} & 2019 & 0.0531 & 0.0486 & 0.0495 & 0.0525 & 0.0470 & 0.0493 & 0.0459 & 0.0618 & 0.0588 & 0.0527 & 0.0521 & 0.0489 & 0.0522 & 0.0450 & 0.0475 & 0.0437 & 0.1131 & 0.0410 & 0.0418 \\
~ & 2020 & 0.0521 & 0.0502 & 0.0517 & 0.0543 & 0.0485 & 0.0541 & 0.0514 & 0.0594 & 0.0600 & 0.0569 & 0.0565 & 0.0507 & 0.0522 & 0.0435 & 0.0495 & 0.0451 & 0.0951 & 0.0331 & 0.0344 \\
~ & 2021 & 0.0471 & 0.0481 & 0.0469 & 0.0516 & 0.0441 & 0.0468 & 0.0445 & 0.1017 & 0.0590 & 0.0508 & 0.0523 & 0.0457 & 0.0495 & 0.0620 & 0.0448 & 0.0415 & 0.1030 & 0.0501 & 0.0536 \\
~ & 2022 & 0.0520 & 0.0522 & 0.0513 & 0.0553 & 0.0485 & 0.0517 & 0.0493 & 0.0732 & 0.0617 & 0.0558 & 0.0573 & 0.0517 & 0.0549 & 0.0466 & 0.0414 & 0.0411 & 0.1205 & 0.0430 & 0.0449 \\ \hline
\multirow{4}{*}{ret\_eom} & 2019 & 0.0360 & 0.0203 & 0.0211 & 0.0305 & 0.0202 & 0.0245 & 0.0195 & 0.0312 & 0.0186 & 0.0184 & 0.0185 & 0.0183 & 0.0142 & 0.0178 & 0.0177 & 0.0229 & 1.0074 & 0.0241 & 0.0240 \\
~ & 2020 & 0.0392 & 0.0193 & 0.0210 & 0.0315 & 0.0194 & 0.0246 & 0.0214 & 0.0411 & 0.0178 & 0.0178 & 0.0178 & 0.0176 & 0.0135 & 0.0170 & 0.0173 & 0.0215 & 1.4456 & 0.0232 & 0.0225 \\
~ & 2021 & 0.0298 & 0.0201 & 0.0210 & 0.0307 & 0.0204 & 0.0217 & 0.0191 & 0.0374 & 0.0187 & 0.0187 & 0.0187 & 0.0185 & 0.0150 & 0.0182 & 0.0178 & 0.0233 & 3.6173 & 0.0230 & 0.0226 \\
~ & 2022 & 0.0314 & 0.0195 & 0.0211 & 0.0279 & 0.0195 & 0.0206 & 0.0185 & 0.0287 & 0.0183 & 0.0187 & 0.0187 & 0.0185 & 0.0148 & 0.0185 & 0.0180 & 0.0217 & 1.9141 & 0.0483 & 0.0302 \\ \hline
\multirow{4}{*}{ret\_l5m} & 2019 & 0.0337 & 0.0201 & 0.0208 & 0.0222 & 0.0197 & 0.0204 & 0.0194 & 0.0295 & 0.0191 & 0.0242 & 0.0200 & 0.0184 & 0.0166 & 0.0192 & 0.0192 & 0.0214 & 0.0827 & 0.0178 & 0.0209 \\
~ & 2020 & 0.0342 & 0.0191 & 0.0207 & 0.0224 & 0.0193 & 0.0206 & 0.0197 & 0.0398 & 0.0188 & 0.0239 & 0.0205 & 0.0177 & 0.0163 & 0.0181 & 0.0184 & 0.0200 & 0.0898 & 0.0171 & 0.0195 \\
~ & 2021 & 0.0279 & 0.0199 & 0.0208 & 0.0222 & 0.0195 & 0.0199 & 0.0190 & 0.0360 & 0.0194 & 0.0253 & 0.0196 & 0.0184 & 0.0174 & 0.0194 & 0.0194 & 0.0215 & 0.0638 & 0.0217 & 0.0213 \\
~ & 2022 & 0.0304 & 0.0194 & 0.0206 & 0.0214 & 0.0185 & 0.0193 & 0.0185 & 0.0287 & 0.0198 & 0.0252 & 0.0196 & 0.0185 & 0.0182 & 0.0195 & 0.0197 & 0.0211 & 0.0630 & 0.0349 & 0.0247 \\ \hline
\multirow{4}{*}{ret\_ldm} & 2019 & 0.0342 & 0.0213 & 0.0208 & 0.0218 & 0.0192 & 0.0200 & 0.0190 & 0.0315 & 0.0198 & 0.0220 & 0.0197 & 0.0185 & 0.0176 & 0.0191 & 0.0195 & 0.0238 & 0.1585 & 0.0213 & 0.0225 \\
~ & 2020 & 0.0345 & 0.0203 & 0.0208 & 0.0221 & 0.0188 & 0.0203 & 0.0195 & 0.0425 & 0.0197 & 0.0222 & 0.0205 & 0.0178 & 0.0171 & 0.0179 & 0.0186 & 0.0221 & 0.1601 & 0.0213 & 0.0212 \\
~ & 2021 & 0.0280 & 0.0208 & 0.0206 & 0.0216 & 0.0189 & 0.0194 & 0.0187 & 0.0381 & 0.0196 & 0.0228 & 0.0194 & 0.0186 & 0.0182 & 0.0190 & 0.0195 & 0.0241 & 0.1118 & 0.0305 & 0.0235 \\
~ & 2022 & 0.0304 & 0.0200 & 0.0204 & 0.0206 & 0.0181 & 0.0184 & 0.0177 & 0.0292 & 0.0201 & 0.0232 & 0.0192 & 0.0185 & 0.0192 & 0.0190 & 0.0200 & 0.0226 & 0.1171 & 0.0450 & 0.0264 \\ \hline
\multirow{4}{*}{fut\_3d} & 2019 & 0.0214 & 0.0210 & 0.0250 & 0.0235 & 0.0152 & 0.0195 & 0.0138 & 0.0253 & 0.0258 & 0.0322 & 0.0304 & 0.0247 & 0.0205 & 0.0125 & 0.0123 & 0.0184 & 1.3412 & 0.0101 & 0.0075 \\
~ & 2020 & 0.0259 & 0.0246 & 0.0315 & 0.0284 & 0.0201 & 0.0250 & 0.0198 & 0.0422 & 0.0305 & 0.0368 & 0.0335 & 0.0298 & 0.0261 & 0.0178 & 0.0163 & 0.0227 & 48.0640 & 0.0142 & 0.0107 \\
~ & 2021 & 0.0226 & 0.0211 & 0.0248 & 0.0256 & 0.0162 & 0.0217 & 0.0141 & 0.0316 & 0.0241 & 0.0289 & 0.0298 & 0.0234 & 0.0218 & 0.0135 & 0.0141 & 0.0177 & 0.1583 & 0.0116 & 0.0084 \\
~ & 2022 & 0.0221 & 0.0194 & 0.0234 & 0.0235 & 0.0160 & 0.0193 & 0.0140 & 0.0230 & 0.0241 & 0.0296 & 0.0293 & 0.0229 & 0.0201 & 0.0130 & 0.0128 & 0.0178 & 37.2985 & 0.0101 & 0.0090 \\ \hline
\multirow{4}{*}{fut\_7d} & 2019 & 0.0215 & 0.0186 & 0.0225 & 0.0213 & 0.0143 & 0.0175 & 0.0141 & 0.0251 & 0.0264 & 0.0768 & 0.0314 & 0.0249 & 0.0227 & 0.0134 & 0.0092 & 0.0188 & 0.6125 & 0.0091 & 0.0076 \\
~ & 2020 & 0.0266 & 0.0219 & 0.0283 & 0.0255 & 0.0181 & 0.0225 & 0.0193 & 0.0409 & 0.0305 & 0.0805 & 0.0345 & 0.0298 & 0.0286 & 0.0183 & 0.0124 & 0.0233 & 0.8126 & 0.0125 & 0.0104 \\
~ & 2021 & 0.0233 & 0.0185 & 0.0222 & 0.0227 & 0.0149 & 0.0190 & 0.0141 & 0.0311 & 0.0246 & 0.0682 & 0.0306 & 0.0236 & 0.0233 & 0.0144 & 0.0107 & 0.0177 & 0.6152 & 0.0102 & 0.0083 \\
~ & 2022 & 0.0227 & 0.0171 & 0.0212 & 0.0212 & 0.0156 & 0.0175 & 0.0150 & 0.0222 & 0.0254 & 0.0740 & 0.0307 & 0.0232 & 0.0222 & 0.0139 & 0.0101 & 0.0185 & 0.6271 & 0.0089 & 0.0084 \\ \hline
\multirow{4}{*}{fut\_15d} & 2019 & 0.0218 & 0.0194 & 0.0230 & 0.0205 & 0.0150 & 0.0189 & 0.0260 & 0.0242 & 0.0242 & 0.1293 & 0.0340 & 0.0260 & 0.0252 & 0.0142 & 0.0085 & 0.0186 & 0.6105 & 0.0088 & 0.0076 \\
~ & 2020 & 0.0268 & 0.0225 & 0.0289 & 0.0243 & 0.0187 & 0.0239 & 0.0270 & 0.0382 & 0.0271 & 0.1388 & 0.0371 & 0.0310 & 0.0312 & 0.0191 & 0.0113 & 0.0230 & 0.8037 & 0.0118 & 0.0099 \\
~ & 2021 & 0.0235 & 0.0193 & 0.0228 & 0.0218 & 0.0155 & 0.0206 & 0.0226 & 0.0298 & 0.0223 & 0.1217 & 0.0333 & 0.0247 & 0.0259 & 0.0148 & 0.0097 & 0.0176 & 0.6111 & 0.0096 & 0.0082 \\
~ & 2022 & 0.0231 & 0.0181 & 0.0217 & 0.0206 & 0.0159 & 0.0184 & 0.0259 & 0.0209 & 0.0230 & 0.1271 & 0.0332 & 0.0242 & 0.0246 & 0.0147 & 0.0093 & 0.0185 & 0.6279 & 0.0084 & 0.0083 \\ \hline
\multirow{4}{*}{fut\_30d} & 2019 & 0.0225 & 0.0210 & 0.0242 & 0.0213 & 0.0153 & 0.0543 & 0.1493 & 0.0285 & 0.0258 & 0.2079 & 0.0571 & 0.0281 & 0.0238 & 0.0592 & 0.0175 & 0.0190 & 0.1311 & 0.0096 & 0.0076 \\
~ & 2020 & 0.0278 & 0.0242 & 0.0305 & 0.0252 & 0.0190 & 0.0578 & 0.1485 & 0.0437 & 0.0284 & 0.2195 & 0.0607 & 0.0335 & 0.0287 & 0.0578 & 0.0218 & 0.0233 & 0.1396 & 0.0129 & 0.0101 \\
~ & 2021 & 0.0244 & 0.0210 & 0.0240 & 0.0228 & 0.0159 & 0.0579 & 0.1504 & 0.0345 & 0.0239 & 0.1983 & 0.0560 & 0.0267 & 0.0246 & 0.0593 & 0.0188 & 0.0180 & 0.1310 & 0.0106 & 0.0084 \\
~ & 2022 & 0.0237 & 0.0197 & 0.0231 & 0.0214 & 0.0163 & 0.0597 & 0.1501 & 0.0256 & 0.0245 & 0.2000 & 0.0553 & 0.0263 & 0.0229 & 0.0579 & 0.0181 & 0.0188 & 0.1337 & 0.0093 & 0.0083 \\ \hline
\multirow{4}{*}{past\_3d} & 2019 & 0.0193 & 0.0210 & 0.0315 & 0.0224 & 0.0174 & 0.0179 & 0.0214 & 0.0322 & 0.0396 & 0.1118 & 0.0328 & 0.0148 & 0.0131 & 0.0118 & 0.0121 & 0.0223 & 0.1970 & 0.0163 & 0.0151 \\
~ & 2020 & 0.0234 & 0.0242 & 0.0401 & 0.0270 & 0.0206 & 0.0210 & 0.0257 & 0.0543 & 0.0435 & 0.1207 & 0.0391 & 0.0176 & 0.0153 & 0.0139 & 0.0145 & 0.0270 & 0.2201 & 0.0205 & 0.0175 \\
~ & 2021 & 0.0198 & 0.0204 & 0.0306 & 0.0216 & 0.0167 & 0.0170 & 0.0209 & 0.0416 & 0.0371 & 0.1030 & 0.0316 & 0.0140 & 0.0123 & 0.0109 & 0.0115 & 0.0213 & 0.1517 & 0.0166 & 0.0142 \\
~ & 2022 & 0.0197 & 0.0195 & 0.0301 & 0.0205 & 0.0162 & 0.0161 & 0.0197 & 0.0308 & 0.0369 & 0.1026 & 0.0313 & 0.0136 & 0.0118 & 0.0104 & 0.0111 & 0.0211 & 0.1604 & 0.0154 & 0.0140 \\ \hline
\multirow{4}{*}{past\_7d} & 2019 & 0.0192 & 0.0204 & 0.0330 & 0.0210 & 0.0151 & 0.0138 & 0.0284 & 0.0351 & 0.0514 & 0.2520 & 0.0406 & 0.0141 & 0.0166 & 0.0144 & 0.0113 & 0.0221 & 0.6559 & 0.0196 & 0.0139 \\
~ & 2020 & 0.0231 & 0.0234 & 0.0416 & 0.0253 & 0.0175 & 0.0160 & 0.0326 & 0.0581 & 0.0558 & 0.2721 & 0.0468 & 0.0166 & 0.0191 & 0.0166 & 0.0133 & 0.0266 & 0.8647 & 0.0257 & 0.0161 \\
~ & 2021 & 0.0199 & 0.0196 & 0.0323 & 0.0202 & 0.0145 & 0.0131 & 0.0275 & 0.0454 & 0.0487 & 0.2372 & 0.0392 & 0.0133 & 0.0157 & 0.0133 & 0.0107 & 0.0210 & 0.6622 & 0.0192 & 0.0131 \\
~ & 2022 & 0.0199 & 0.0190 & 0.0318 & 0.0192 & 0.0141 & 0.0123 & 0.0266 & 0.0322 & 0.0477 & 0.2379 & 0.0388 & 0.0129 & 0.0152 & 0.0129 & 0.0103 & 0.0206 & 0.6784 & 0.0183 & 0.0127 \\ \hline
\multirow{4}{*}{past\_15d} & 2019 & 0.0195 & 0.0204 & 0.0327 & 0.0211 & 0.0149 & 0.0157 & 0.0665 & 0.0396 & 0.0494 & 0.2804 & 0.0537 & 0.0137 & 0.0178 & 0.0340 & 0.0117 & 0.0211 & 0.6970 & 0.0381 & 0.0124 \\
~ & 2020 & 0.0234 & 0.0234 & 0.0410 & 0.0255 & 0.0171 & 0.0182 & 0.0717 & 0.0645 & 0.0531 & 0.3013 & 0.0605 & 0.0161 & 0.0202 & 0.0374 & 0.0136 & 0.0253 & 0.9080 & 0.0491 & 0.0144 \\
~ & 2021 & 0.0203 & 0.0197 & 0.0321 & 0.0201 & 0.0143 & 0.0150 & 0.0650 & 0.0504 & 0.0463 & 0.2662 & 0.0522 & 0.0129 & 0.0167 & 0.0328 & 0.0110 & 0.0201 & 0.7022 & 0.0378 & 0.0118 \\
~ & 2022 & 0.0204 & 0.0189 & 0.0317 & 0.0189 & 0.0138 & 0.0141 & 0.0641 & 0.0381 & 0.0448 & 0.2667 & 0.0519 & 0.0125 & 0.0162 & 0.0320 & 0.0106 & 0.0199 & 0.7240 & 0.0360 & 0.0113 \\ \hline
\multirow{4}{*}{past\_30d} & 2019 & 0.0214 & 0.0210 & 0.0354 & 0.0355 & 0.0185 & 0.1202 & 0.2118 & 0.0496 & 0.0573 & 0.2882 & 0.1741 & 0.0244 & 0.0245 & 0.1068 & 0.0594 & 0.0237 & 1.0544 & 0.0706 & 0.0119 \\
~ & 2020 & 0.0257 & 0.0241 & 0.0441 & 0.0416 & 0.0213 & 0.1246 & 0.2173 & 0.0792 & 0.0617 & 0.3090 & 0.1834 & 0.0287 & 0.0278 & 0.1118 & 0.0673 & 0.0282 & 1.3091 & 0.0874 & 0.0140 \\
~ & 2021 & 0.0221 & 0.0203 & 0.0350 & 0.0334 & 0.0177 & 0.1180 & 0.2100 & 0.0643 & 0.0543 & 0.2755 & 0.1718 & 0.0230 & 0.0230 & 0.1051 & 0.0577 & 0.0223 & 1.0560 & 0.0709 & 0.0113 \\
~ & 2022 & 0.0221 & 0.0194 & 0.0350 & 0.0321 & 0.0170 & 0.1166 & 0.2088 & 0.0489 & 0.0525 & 0.2762 & 0.1718 & 0.0222 & 0.0221 & 0.1037 & 0.0563 & 0.0221 & 1.0838 & 0.0679 & 0.0109 \\
\bottomrule
\bottomrule
\end{tabular}}
\caption{Annual MAE Trends (2019–2022) on the MM-FinEval Benchmark.}
\label{tab:MAE_trend}
\end{table*}

\section{Examples of Model Responses}
\label{app:example}

\begin{tcolorbox}[
    enhanced,
    breakable,
    colback=blue!8!white, % Light blue/violet background
    colframe=blue!50!black, % Dark blue frame
    arc=2mm,
    boxrule=1pt,
    title={\bfseries Qwen3.5 2B},
    coltitle=white,
    attach boxed title to top left={yshift=-2mm, xshift=3mm},
    boxed title style={enhanced, colback=blue!50!black, colframe=blue!50!black, arc=2mm, boxrule=0pt},
    top=0.5mm, 
    left=1mm, 
    right=1mm, 
    bottom=0.5mm,
]
\setlength{\parskip}{2pt}
\small\vskip8pt

Rationale = The Cumulative Abnormal Return (CAR) is predicted to be positive, driven by the company's achievement of all-time record quarterly and full-year operating income and operating ratio, which significantly outperformed analyst expectations and the broader market.

Prediction = [[ 0.045 ]]

\end{tcolorbox}

\begin{tcolorbox}[
    enhanced,
    breakable,
    colback=blue!8!white, % Light blue/violet background
    colframe=blue!50!black, % Dark blue frame
    arc=2mm,
    boxrule=1pt,
    title={\bfseries Qwen3.5 4B},
    coltitle=white,
    attach boxed title to top left={yshift=-2mm, xshift=3mm},
    boxed title style={enhanced, colback=blue!50!black, colframe=blue!50!black, arc=2mm, boxrule=0pt},
    top=0.5mm, 
    left=1mm, 
    right=1mm, 
    bottom=0.5mm,
]
\setlength{\parskip}{2pt}
\small\vskip8pt

Rationale = [[ The company reported a 22\% decline in GAAP earnings per share due to significant one-time costs, including severance charges and litigation expenses, which negatively impacted the market's sentiment despite underlying non-GAAP growth. The management's focus on cost-cutting and the uncertainty surrounding future market conditions likely led to a negative market reaction.]]

Prediction = [[ -0.035 ]]

\end{tcolorbox}

\begin{tcolorbox}[
    enhanced,
    breakable,
    colback=blue!8!white, % Light blue/violet background
    colframe=blue!50!black, % Dark blue frame
    arc=2mm,
    boxrule=1pt,
    title={\bfseries Qwen3.5 9B},
    coltitle=white,
    attach boxed title to top left={yshift=-2mm, xshift=3mm},
    boxed title style={enhanced, colback=blue!50!black, colframe=blue!50!black, arc=2mm, boxrule=0pt},
    top=0.5mm, 
    left=1mm, 
    right=1mm, 
    bottom=0.5mm,
]
\setlength{\parskip}{2pt}
\small\vskip8pt

Rationale = The company delivered solid full-year results with revenue growth, margin expansion, and strong free cash flow, but the stock price likely already priced in the positive momentum from the second half of the year. Additionally, the guidance for 2019 EPS of \$13.90 was presented as a \"floor\" rather than a target, and the company explicitly noted that the Red Hat acquisition (a major value driver) is excluded from 2019 financial expectations due to timing uncertainty, which may limit the upside surprise in the immediate reaction.

Prediction = [[ 0.012 ]]

\end{tcolorbox}

\begin{tcolorbox}[
    enhanced,
    breakable,
    colback=blue!8!white, % Light blue/violet background
    colframe=blue!50!black, % Dark blue frame
    arc=2mm,
    boxrule=1pt,
    title={\bfseries Qwen3.5 27B},
    coltitle=white,
    attach boxed title to top left={yshift=-2mm, xshift=3mm},
    boxed title style={enhanced, colback=blue!50!black, colframe=blue!50!black, arc=2mm, boxrule=0pt},
    top=0.5mm, 
    left=1mm, 
    right=1mm, 
    bottom=0.5mm,
]
\setlength{\parskip}{2pt}
\small\vskip8pt

Rationale = [[ The company reported a 22\% year-over-year decline in GAAP earnings per share to \$0.84, missing expectations due to significant one-time charges for severance, litigation, and headquarters relocation totaling \$0.16 per share. Although non-GAAP earnings grew 9\%, the market reaction is likely to be negative due to the headline GAAP miss and the explicit guidance that equity markets will not be a significant contributor to revenue growth in 2019. ]]

Prediction = [[ -0.035 ]]

\end{tcolorbox}

\begin{tcolorbox}[
    enhanced,
    breakable,
    colback=blue!8!white, % Light blue/violet background
    colframe=blue!50!black, % Dark blue frame
    arc=2mm,
    boxrule=1pt,
    title={\bfseries Qwen3.5 35B},
    coltitle=white,
    attach boxed title to top left={yshift=-2mm, xshift=3mm},
    boxed title style={enhanced, colback=blue!50!black, colframe=blue!50!black, arc=2mm, boxrule=0pt},
    top=0.5mm, 
    left=1mm, 
    right=1mm, 
    bottom=0.5mm,
]
\setlength{\parskip}{2pt}
\small\vskip8pt

Rationale = PNC delivered a strong 2020 with record revenue and positive operating leverage, but the market reaction is likely tempered by the guidance for a modest decline in Net Interest Income (NII) for 2021 and the significant drag on reserves (\$3.2 billion provision) due to the pandemic, resulting in a neutral to slightly negative market response.

Prediction = -0.012

\end{tcolorbox}

\begin{tcolorbox}[
    enhanced,
    breakable,
    colback=blue!8!white, % Light blue/violet background
    colframe=blue!50!black, % Dark blue frame
    arc=2mm,
    boxrule=1pt,
    title={\bfseries Qwen3.6 27B},
    coltitle=white,
    attach boxed title to top left={yshift=-2mm, xshift=3mm},
    boxed title style={enhanced, colback=blue!50!black, colframe=blue!50!black, arc=2mm, boxrule=0pt},
    top=0.5mm, 
    left=1mm, 
    right=1mm, 
    bottom=0.5mm,
]
\setlength{\parskip}{2pt}
\small\vskip8pt

Rationale = [[ The company significantly raised its full-year 2021 revenue guidance to a range of \$5.5-\$6.0 billion and projected a substantial expansion in operating margins to 10.4\%-11.0\%, driven by strong net order growth and a robust backlog. Additionally, the firm highlighted a strong balance sheet with a low debt-to-capital ratio and a clear path to double-digit community count growth, signaling strong future profitability. ]]

Prediction = [[ 0.065 ]]

\end{tcolorbox}

\begin{tcolorbox}[
    enhanced,
    breakable,
    colback=blue!8!white, % Light blue/violet background
    colframe=blue!50!black, % Dark blue frame
    arc=2mm,
    boxrule=1pt,
    title={\bfseries Qwen3.6 35B},
    coltitle=white,
    attach boxed title to top left={yshift=-2mm, xshift=3mm},
    boxed title style={enhanced, colback=blue!50!black, colframe=blue!50!black, arc=2mm, boxrule=0pt},
    top=0.5mm, 
    left=1mm, 
    right=1mm, 
    bottom=0.5mm,
]
\setlength{\parskip}{2pt}
\small\vskip8pt

Rationale = BlackRock reported record full-year revenue growth of 20\% and operating income growth of 19\%, significantly outperforming expectations with record net inflows of \$540 billion and a 10\% increase in AUM to \$10 trillion. The company also announced a strong capital return program, including an 18\% dividend increase and \$1.5 billion in share repurchases, signaling high confidence in future cash flows and growth.

Prediction = [[ 0.035 ]]

\end{tcolorbox}

\begin{tcolorbox}[
    enhanced,
    breakable,
    colback=blue!8!white, % Light blue/violet background
    colframe=blue!50!black, % Dark blue frame
    arc=2mm,
    boxrule=1pt,
    title={\bfseries Gemma3n},
    coltitle=white,
    attach boxed title to top left={yshift=-2mm, xshift=3mm},
    boxed title style={enhanced, colback=blue!50!black, colframe=blue!50!black, arc=2mm, boxrule=0pt},
    top=0.5mm, 
    left=1mm, 
    right=1mm, 
    bottom=0.5mm,
]
\setlength{\parskip}{2pt}
\small\vskip8pt

Rationale = [[ The earnings call presented a strong Q3 performance with solid growth across key products like Opdivo and Eliquis, and positive momentum from new launches like Zeposia and Abecma. The company reaffirmed its financial guidance and expressed confidence in its ability to navigate the challenges of generic erosion while continuing to invest in innovation, which is likely to be viewed favorably by investors. ]]

Prediction = [[ 0.06 ]]

\end{tcolorbox}

\begin{tcolorbox}[
    enhanced,
    breakable,
    colback=blue!8!white, % Light blue/violet background
    colframe=blue!50!black, % Dark blue frame
    arc=2mm,
    boxrule=1pt,
    title={\bfseries Mistral3 3B},
    coltitle=white,
    attach boxed title to top left={yshift=-2mm, xshift=3mm},
    boxed title style={enhanced, colback=blue!50!black, colframe=blue!50!black, arc=2mm, boxrule=0pt},
    top=0.5mm, 
    left=1mm, 
    right=1mm, 
    bottom=0.5mm,
]
\setlength{\parskip}{2pt}
\small\vskip8pt

Rationale = [The strong execution in 2018, with significant improvements in ROTCE (14.1\%), efficiency ratio (56.7\%), and underlying EPS growth, alongside disciplined balance sheet growth and robust credit quality, suggests strong investor confidence. The company's strategic initiatives like TOP programs, balance sheet optimization, and digital transformation, combined with a positive economic outlook and prudent risk management, should drive positive abnormal returns. The stock's current valuation discount relative to peers, especially in a recession-sensitive sector, and the company's ability to sustain growth through organic and strategic acquisitions, are likely to support a positive CAR.]

Prediction = [0.065]

\end{tcolorbox}

\begin{tcolorbox}[
    enhanced,
    breakable,
    colback=blue!8!white, % Light blue/violet background
    colframe=blue!50!black, % Dark blue frame
    arc=2mm,
    boxrule=1pt,
    title={\bfseries Mistral3 8B},
    coltitle=white,
    attach boxed title to top left={yshift=-2mm, xshift=3mm},
    boxed title style={enhanced, colback=blue!50!black, colframe=blue!50!black, arc=2mm, boxrule=0pt},
    top=0.5mm, 
    left=1mm, 
    right=1mm, 
    bottom=0.5mm,
]
\setlength{\parskip}{2pt}
\small\vskip8pt

Rationale =Capital One's fourth quarter 2019 earnings call and presentation highlighted several positive yet nuanced financial developments. The company reported strong loan growth (7\% year-over-year), solid revenue growth (7\%), and a stable credit environment, though offset by increased provisions for credit losses (up 11\%) and the transition to CECL, which added \$2.9B to the allowance. The Walmart partnership launch, digital transformation progress, and moderate marketing expense growth suggest sustained growth momentum, but the CECL impact and higher provisioning could dampen near-term profitability expectations. The net charge-off rate remained stable at 1.93\%, and the company's disciplined capital management and strategic investments in technology and marketing indicate long-term value creation.

Prediction = [[ 0.025 ]]",

\end{tcolorbox}

\begin{tcolorbox}[
    enhanced,
    breakable,
    colback=blue!8!white, % Light blue/violet background
    colframe=blue!50!black, % Dark blue frame
    arc=2mm,
    boxrule=1pt,
    title={\bfseries Mistral3 14B},
    coltitle=white,
    attach boxed title to top left={yshift=-2mm, xshift=3mm},
    boxed title style={enhanced, colback=blue!50!black, colframe=blue!50!black, arc=2mm, boxrule=0pt},
    top=0.5mm, 
    left=1mm, 
    right=1mm, 
    bottom=0.5mm,
]
\setlength{\parskip}{2pt}
\small\vskip8pt

Rationale = [[BNY Mellon's earnings call highlighted strong organic growth, innovation in digital assets and real-time payments, and a robust capital position, which should positively surprise investors. The guidance for double-digit EPS growth in 2022, driven by higher net interest revenue and fee revenue recovery, further supports a positive market reaction.]]

Prediction = [[0.035]]

\end{tcolorbox}

\begin{tcolorbox}[
    enhanced,
    breakable,
    colback=blue!8!white, % Light blue/violet background
    colframe=blue!50!black, % Dark blue frame
    arc=2mm,
    boxrule=1pt,
    title={\bfseries GPT 5.4},
    coltitle=white,
    attach boxed title to top left={yshift=-2mm, xshift=3mm},
    boxed title style={enhanced, colback=blue!50!black, colframe=blue!50!black, arc=2mm, boxrule=0pt},
    top=0.5mm, 
    left=1mm, 
    right=1mm, 
    bottom=0.5mm,
]
\setlength{\parskip}{2pt}
\small\vskip8pt

Rationale = [[ Kimberly-Clark delivered a clear 2020 beat with strong EPS, margin expansion, cash flow, and improved market share, but the market reaction is likely tempered by a cautious 2021 outlook featuring only 1-2\% organic growth and significant commodity inflation that is expected to pressure margins. The net effect should be modestly positive because execution and restructuring/cost savings were strong, yet guidance implies a tougher earnings quality and margin backdrop ahead. ]]

Prediction = [[ 0.012 ]]

\end{tcolorbox}

\begin{tcolorbox}[
    enhanced,
    breakable,
    colback=blue!8!white, % Light blue/violet background
    colframe=blue!50!black, % Dark blue frame
    arc=2mm,
    boxrule=1pt,
    title={\bfseries GPT 5.5},
    coltitle=white,
    attach boxed title to top left={yshift=-2mm, xshift=3mm},
    boxed title style={enhanced, colback=blue!50!black, colframe=blue!50!black, arc=2mm, boxrule=0pt},
    top=0.5mm, 
    left=1mm, 
    right=1mm, 
    bottom=0.5mm,
]
\setlength{\parskip}{2pt}
\small\vskip8pt

Rationale = [[ Despite record full-year results, the event was dominated by investor concern over a Q4 EPS miss driven by sharply higher compensation/non-comp expenses, wage inflation, and weaker sequential trading performance. Management's commentary on continued investment spending, capital headwinds, and limited near-term buybacks likely reinforced negative revisions to forward profitability. ]]

Prediction = [[ -0.065 ]]

\end{tcolorbox}

\begin{tcolorbox}[
    enhanced,
    breakable,
    colback=blue!8!white, % Light blue/violet background
    colframe=blue!50!black, % Dark blue frame
    arc=2mm,
    boxrule=1pt,
    title={\bfseries Claude 4.6},
    coltitle=white,
    attach boxed title to top left={yshift=-2mm, xshift=3mm},
    boxed title style={enhanced, colback=blue!50!black, colframe=blue!50!black, arc=2mm, boxrule=0pt},
    top=0.5mm, 
    left=1mm, 
    right=1mm, 
    bottom=0.5mm,
]
\setlength{\parskip}{2pt}
\small\vskip8pt

Rationale = [[ PMI delivered a strong beat with Q2 adjusted diluted EPS of \$1.57 (+17.8\% organic), raised full-year EPS guidance to \$5.97-\$6.07 (implying +15-17\% in dollar terms), confirmed a \$5-7B share buyback program, and announced the imminent IQOS ILUMA Japan launch all meaningfully positive catalysts that likely exceeded consensus expectations and drove a positive abnormal return. ]]

Prediction = [[ 0.038 ]]

\end{tcolorbox}

\begin{tcolorbox}[
    enhanced,
    breakable,
    colback=blue!8!white, % Light blue/violet background
    colframe=blue!50!black, % Dark blue frame
    arc=2mm,
    boxrule=1pt,
    title={\bfseries Claude 4.8},
    coltitle=white,
    attach boxed title to top left={yshift=-2mm, xshift=3mm},
    boxed title style={enhanced, colback=blue!50!black, colframe=blue!50!black, arc=2mm, boxrule=0pt},
    top=0.5mm, 
    left=1mm, 
    right=1mm, 
    bottom=0.5mm,
]
\setlength{\parskip}{2pt}
\small\vskip8pt

Rationale = [[ WBA delivered a strong Q1 beat (adjusted EPS \$1.68, up 53\%, well ahead of expectations) driven by COVID vaccinations/testing and robust retail comps, plus raised full-year guidance, which should be received positively despite headwinds from shrink, labor investments, and softer underlying scripts. ]]

Prediction = [[ 0.045 ]]

\end{tcolorbox}

\begin{tcolorbox}[
    enhanced,
    breakable,
    colback=blue!8!white, % Light blue/violet background
    colframe=blue!50!black, % Dark blue frame
    arc=2mm,
    boxrule=1pt,
    title={\bfseries Voxtral 24B},
    coltitle=white,
    attach boxed title to top left={yshift=-2mm, xshift=3mm},
    boxed title style={enhanced, colback=blue!50!black, colframe=blue!50!black, arc=2mm, boxrule=0pt},
    top=0.5mm, 
    left=1mm, 
    right=1mm, 
    bottom=0.5mm,
]
\setlength{\parskip}{2pt}
\small\vskip8pt

**Answer**

Rationale = The company's strong financial performance, with a 32\% increase in comparable basis EPS excluding canopy equity earnings, and the announcement of a \$2 billion share repurchase program, are the key drivers for the stock's CAR. The company's commitment to returning \$5 billion in value to shareholders through fiscal 23 also supports a positive outlook.

Prediction = 0.035

\end{tcolorbox}

\begin{tcolorbox}[
    enhanced,
    breakable,
    colback=blue!8!white, % Light blue/violet background
    colframe=blue!50!black, % Dark blue frame
    arc=2mm,
    boxrule=1pt,
    title={\bfseries Moss},
    coltitle=white,
    attach boxed title to top left={yshift=-2mm, xshift=3mm},
    boxed title style={enhanced, colback=blue!50!black, colframe=blue!50!black, arc=2mm, boxrule=0pt},
    top=0.5mm, 
    left=1mm, 
    right=1mm, 
    bottom=0.5mm,
]
\setlength{\parskip}{2pt}
\small\vskip8pt

Rationale = [[ This earnings call seems to be a 'surprising positive' event which can lead to a strong positive CAR. It may provide a signal that the outlook for the company is improving, leading to an increase in the company's stock price. ]]

Prediction = 0.123

\end{tcolorbox}

\begin{tcolorbox}[
    enhanced,
    breakable,
    colback=blue!8!white, % Light blue/violet background
    colframe=blue!50!black, % Dark blue frame
    arc=2mm,
    boxrule=1pt,
    title={\bfseries Gemma4 E2B},
    coltitle=white,
    attach boxed title to top left={yshift=-2mm, xshift=3mm},
    boxed title style={enhanced, colback=blue!50!black, colframe=blue!50!black, arc=2mm, boxrule=0pt},
    top=0.5mm, 
    left=1mm, 
    right=1mm, 
    bottom=0.5mm,
]
\setlength{\parskip}{2pt}
\small\vskip8pt

Rationale = [[ The company reported strong retail sales growth in Q3, driven by improved mix and digital sales, and provided guidance indicating a strong Q3 and a cautious outlook for Q4. The financial results show a positive trend in retail sales and margin expansion, which are positive indicators for the stock. ]]

Prediction = [[ 0.025 ]]

\end{tcolorbox}

\begin{tcolorbox}[
    enhanced,
    breakable,
    colback=blue!8!white, % Light blue/violet background
    colframe=blue!50!black, % Dark blue frame
    arc=2mm,
    boxrule=1pt,
    title={\bfseries Gemma4 E4B},
    coltitle=white,
    attach boxed title to top left={yshift=-2mm, xshift=3mm},
    boxed title style={enhanced, colback=blue!50!black, colframe=blue!50!black, arc=2mm, boxrule=0pt},
    top=0.5mm, 
    left=1mm, 
    right=1mm, 
    bottom=0.5mm,
]
\setlength{\parskip}{2pt}
\small\vskip8pt

Rationale = [[ The earnings call presents a positive outlook with strong underlying sales growth and a significant increase in adjusted segment EBIT, which should lead to a positive market reaction. The company also highlights strong free cash flow and a positive ESG momentum, which further supports a bullish sentiment. ]]

Prediction = [[ 0.035 ]]

\end{tcolorbox}

\end{document}